\documentclass[final,onefignum,onetabnum]{siamart190516}

\usepackage{lipsum}
\usepackage{amsfonts}
\usepackage{amssymb}
\usepackage{graphicx}
\usepackage{epstopdf}
\usepackage{algorithmic}
\usepackage{enumerate}
\usepackage{verbatim,amscd}
\usepackage{latexsym, bm, float, color}
\usepackage{mathtools}
\usepackage{pifont}
\usepackage{subfig}

\ifpdf
  \DeclareGraphicsExtensions{.eps,.pdf,.png,.jpg}
\else
  \DeclareGraphicsExtensions{.eps}
\fi

\newsiamremark{remark}{Remark}
\newsiamremark{hypothesis}{Hypothesis}
\crefname{hypothesis}{Hypothesis}{Hypotheses}
\newsiamthm{claim}{Claim}

\headers{Plateau Phenomenon}{M. Ainsworth and Y. Shin}

\title{Plateau Phenomenon in Gradient Descent Training of ReLU networks: Explanation, Quantification and Avoidance\thanks{Submitted to the editors 2020/07/14.
\funding{This work was funded by the PhILMS grant DE-SC0019453}}}

\author{Mark Ainsworth\thanks{Division of Applied Mathematics, Brown University, Providence, RI 02912
  (\email{mark\_ainsworth@brown.edu}, \email{yeonjong\_shin@brown.edu}).}
\and Yeonjong Shin\footnotemark[2]}

\usepackage{amsopn}

\ifpdf
\hypersetup{
  pdftitle={Plateau Phenomenon},
  pdfauthor={M. Ainsworth and Y. Shin}
}
\fi

\begin{document}

\maketitle

\begin{abstract}
  The ability of neural networks to provide ‘best in class’ approximation across
  a wide range of applications is well-documented.  Nevertheless, the powerful
  expressivity of neural networks comes to naught if one is unable to effectively
  train (choose) the parameters defining the network. In general, neural networks
  are trained by gradient descent type optimization methods, or a stochastic
  variant thereof.  In practice, such methods result in the loss function
  decreases rapidly at the beginning of training but then, after a relatively
  small number of steps, significantly slow down. The loss may even appear to
  stagnate over the period of a large number of epochs, only to then suddenly
  start to decrease fast again for no apparent reason. This so-called plateau
  phenomenon manifests itself in many learning tasks.
  
  The present work aims to identify and quantify the root causes of plateau
  phenomenon. No assumptions are made on the number of neurons relative to the
  number of training data, and our results hold for both the lazy and adaptive
  regimes. The main findings are: plateaux correspond to periods during which
  activation patterns remain constant, where activation pattern refers to the
  number of data points that activate a given neuron; quantification of
  convergence of the gradient flow dynamics; and, characterization of stationary
  points in terms solutions of local least squares regression lines over subsets
  of the training data. Based on these conclusions, we propose a new iterative
  training method, the Active Neuron Least Squares (ANLS), characterised by the  
  explicit adjustment of the activation pattern at each step, which is designed
  to enable a quick exit from a plateau.  Illustrative numerical examples are
  included throughout.
\end{abstract}




\section{Introduction} \label{sec:intro}
Neural networks are a key component in machine learning \cite{Lecun_Nature15_DeepLearning}.
The ability of neural networks to provide `best in class' approximation 
across a wide range of applications
has been the subject of intensive research \cite{Cybenko_MoC89, Barron_IEEE93, Mhaskar_Neurl96, Pinkus_Acta99, Yarotsky_NN17, Petersen_NN18}.
Nevertheless, the powerful expressivity of neural networks 
comes to naught if one is unable to effectively train (choose)
the parameters defining the network.
In general, neural networks are trained by gradient descent type optimization methods, or a stochastic variant thereof \cite{Ruder_16_GDoverview}.
The compositional structure of networks
combined with nonlinear activation functions
makes the associated optimization problem
non-convex.
Empirical evidence shows that,
despite the non-convex nature of the optimization problem,
such methods often manage to find a good enough solution
for the purposes of practical machine learning applications.
Nevertheless, the convergence of gradient descent methods
is often extremely slow and in many cases the error,
or loss, tends to remain at or above a level of 0.1-1\% relative error,
even when an extremely large number of steps are taken.
The fact that such methods have been
implemented extremely efficiently on state of the art computer hardware
helps ameliorate the problem to a large extent \cite{Abadi_16_TF,Paszke_19Pytorch}.
However, the fact remains that the training of neural networks 
using gradient descent, stochastic gradient descent, \texttt{Adam} \cite{kingma2014adam}, etc.
very probably accounts for more computer cycles than almost any other
application in scientific computing at the present time.

Some investigations have been undertaken
to try to understand the behavior of gradient descent training of neural networks.
Several works \cite{allen2018convergence,du2018gradient-shallow,du2018gradient-DNN,oymak2019towards,zou2018stochastic,Jacot_18NTK} analyzed neural networks
in the so-called \emph{lazy} \cite{Chizat_19_LazyTrain}
(also known as, over-parameterized or kernel) regime,
where the number of network parameters is significantly larger than the number of training data, where it was shown that
gradient descent can train 
neural networks to interpolate all the training data (give a zero training loss).
However, as argued in  \cite{Chizat_19_LazyTrain, Lee_19_WideLinear}, lazy training is akin to simple linear regression type (kernel) learning, is inherently non-unique  and 
can lead to overfitting  \cite{Ghorbani_19_LimitationsLazy}.
In a similar vein \cite{Mei_PNAS18_MFland,Rotskoff_18_NNsInteracting,Sirignano_20_MFANNs} studied neural networks in the mean-field regime, where
the width of networks tends to infinity and, once again, it was shown that under certain conditions, a zero loss can be achieved.
While such limiting cases may be of interest in some machine learning applications, the adaptive regime \cite{Williams_NIPS19GDofShallow},
in which the amount of data exceeds the number of parameters in the model, is generally of more interest in practice.

\begin{figure}[htbp]
	\centerline{
	\subfloat[ ]{\label{fig:moti:a}
	\includegraphics[height=3.3cm]{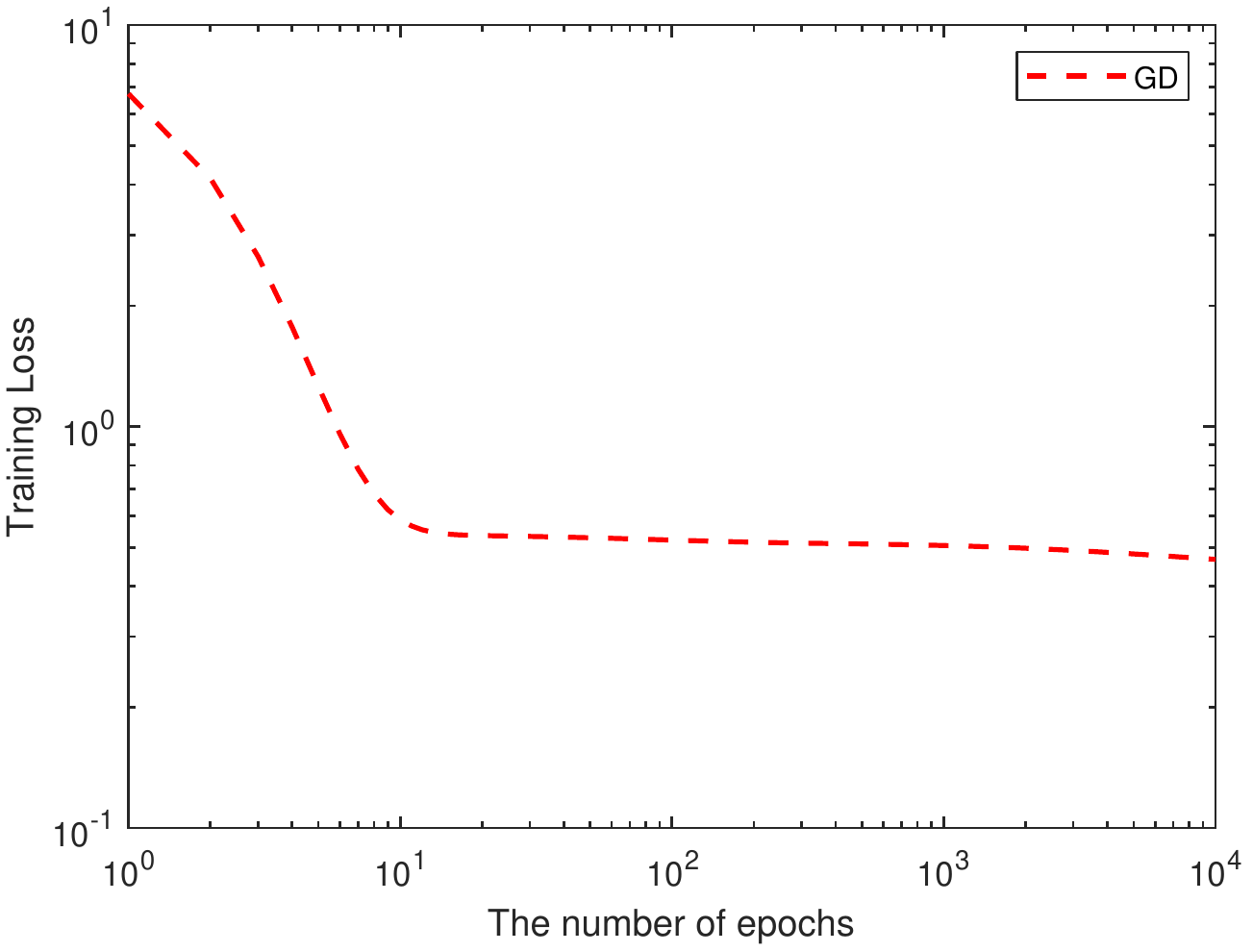}
	}
	\subfloat[ ]{\label{fig:moti:b}
		\includegraphics[height=3.3cm]{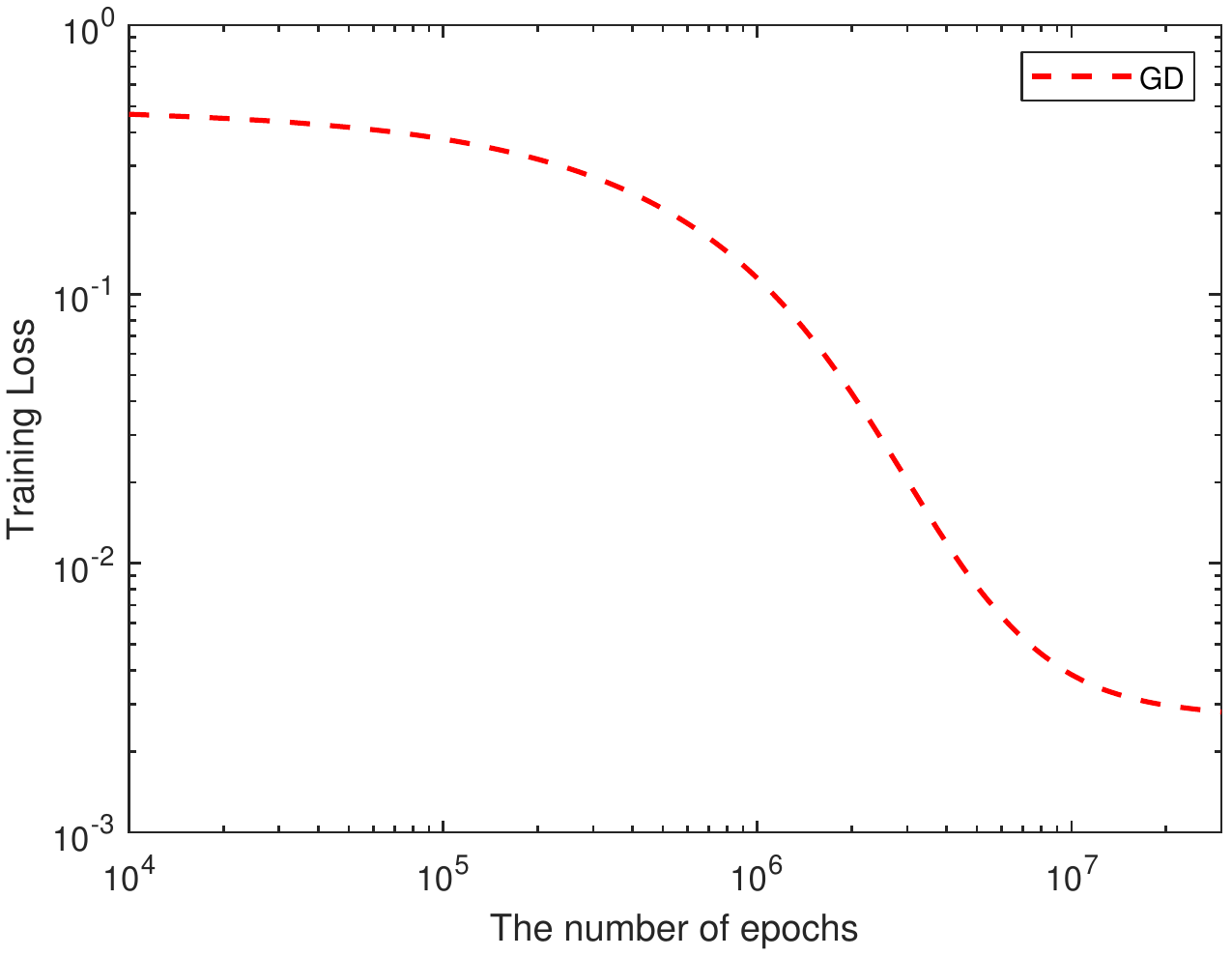}
	}	
	\subfloat[ ]{\label{fig:moti:c}
		\includegraphics[height=3.3cm]{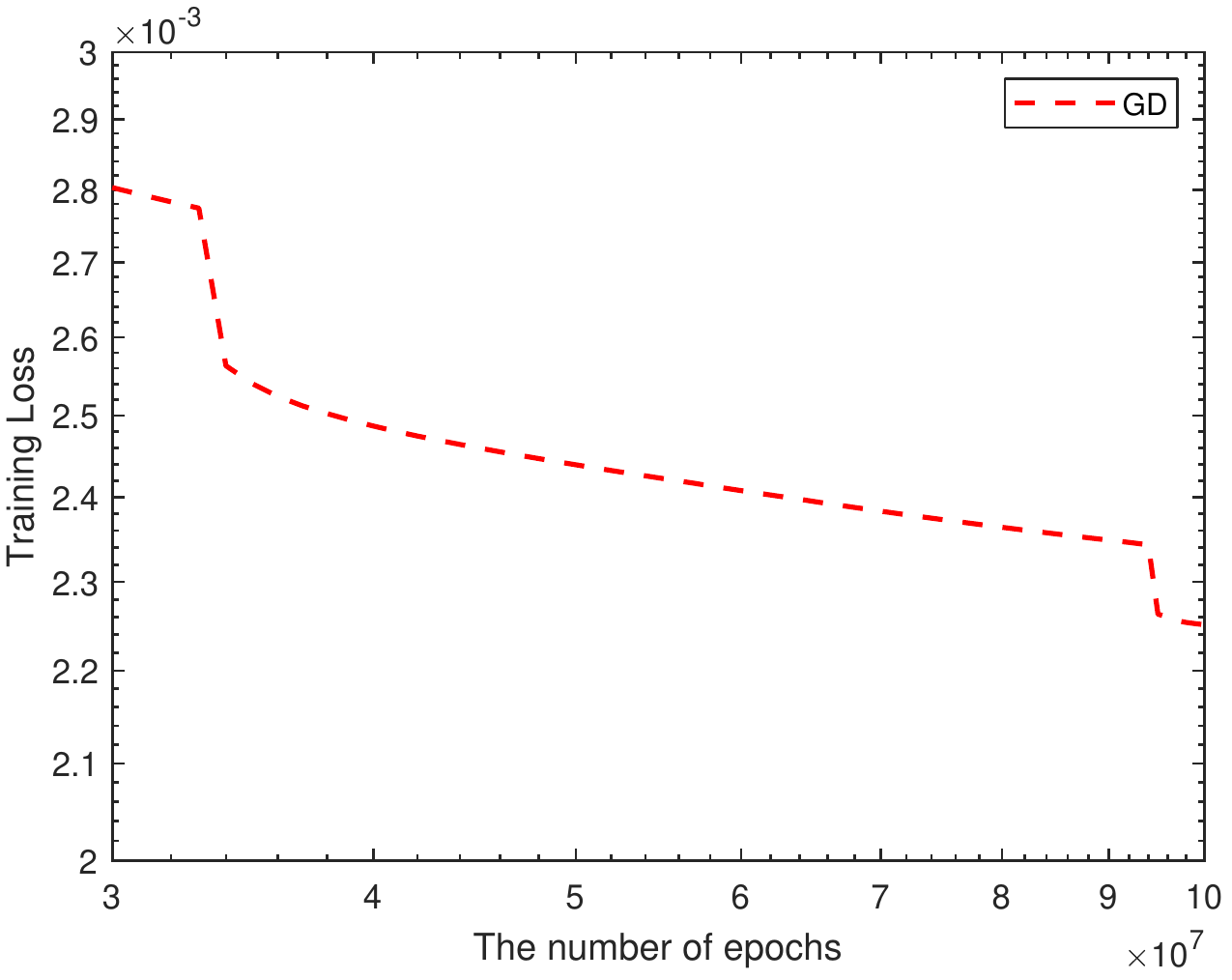}
	}
	}
	\caption{Training loss versus the number of epochs for approximating $\sin(5\pi x)$ with a two-layer ReLU network of width 100 using 
		500 equidistant training data on $[-1,1]$.
		Gradient decent (full-batch) is applied with a learning rate of $10^{-3}$.
	}
	\label{fig:motivation}
\end{figure}

When neural networks belong to the adaptive regime,
we still lack a thorough understanding of gradient descent training of neural networks
and
many questions remain elusive: Will the network converge to a `good' solution? 
Will convergence even occur at all \cite{Montavon_12_NNtricks}?
In practice, it is often observed 
that the loss function decreases rapidly at the beginning of training
but then, after a relatively small number of steps, significantly slows down.
The loss may even appear to stagnate over the period of a large number of epochs, only to then suddenly start to decrease
fast again for no apparent reason. 
This so-called \emph{plateau phenomenon} \cite{Fukumizu_00_Plateaus}
manifests itself in many learning tasks
and is seen even in quite simple applications.
For instance, suppose we wish to approximate $\sin(5\pi x)$ using a two-layer rectified linear unit (ReLU) network of width $100$ over $500$ equidistant training data points on $[-1,1]$.
For the training, we apply the gradient descent on the mean squared loss with a constant learning rate of $10^{-3}$.
In Figure~\ref{fig:motivation}, the loss versus the number of epochs is plotted during different training regimes.
Figure~\ref{fig:moti:a} shows the behavior early in the training in which
the loss decreases quickly but then appears to saturate.
The unseasoned practitioner may even be tempted to terminate the training at this point believing convergence has been achieved.
However, if the gradient decent is continued, as shown in Figure~\ref{fig:moti:b}
the loss suddenly begins to decrease rapidly before again saturating at around $10^7$ epochs.
Continuing the gradient decent further as in Figure~\ref{fig:moti:c}, we observe a similar behavior repeated.

This behavior has been observed by other researchers 
\cite{Saad_PhyRE95, Fukumizu_00_Plateaus} and termed ``plateau phenomenon" \cite{Park_NN00}.
Plateau phenomena not only make it difficult to decide at which stage one should terminate the gradient descent, but also means that convergence is very slow
with increasingly large numbers of seemingly unnecessary iterations 
spent in traversing a plateau only to enter a new regime in which the loss decreases rapidly.
Although \cite{Fukumizu_00_Plateaus,Cousseau_08_dynamics,Wei_08_dynamics,Guo_18_influence,Yoshida_NIPS19Data-Depend-PP}
provided some heuristic and empirical explanations on the cause of the plateau phenomenon,
a rigorous mathematical understanding is still lacking.

One of the aims of the present work is to identify and quantify the root causes of plateau phenomena. 
In order to isolate extraneous effects, we confine our attention to 
univariate two-layer ReLU networks
where, as seen in the above example,
the phenomenon already exhibits itself.
We give a mathematical characterization of
the plateau phenomenon.
No assumptions are made on the number of neurons relative to the number of training data, and our results hold for both the lazy and adaptive regimes.
However, for the reasons mentioned before, 
we are mainly interested in the adaptive regime.

The main findings in the present work are summarized as follows:
\begin{itemize}
	\item 
	Plateaux correspond to periods during which
	\textit{activation patterns} remain constant (Theorem~\ref{thm:conti-loss-n}).
	The term activation pattern is defined below, but roughly speaking, refers to 
	the number of data points that activate a given neuron.
	\item We quantify the convergence of the gradient flow dynamics (Theorem~\ref{thm:convergence});
	give necessary and sufficient conditions for the stationary points (Lemma~\ref{lemma:critical}); and,
	relate fully trained networks to local least squares regression lines (Theorem~\ref{thm:LSQ-Stationary}) over subsets of the training data.
	\item
	Finally, based on these conclusions, 
	we propose a new iterative training method, Active Neuron Least Squares (ANLS), which is designed to avoid 
	the plateau phenomenon altogether (Section~\ref{sec:method}).
\end{itemize}

\begin{figure}[htbp]
	\centerline{
		\subfloat[ ]{\label{fig:moti2:a}
			\includegraphics[width=6.3cm, height=4.5cm]{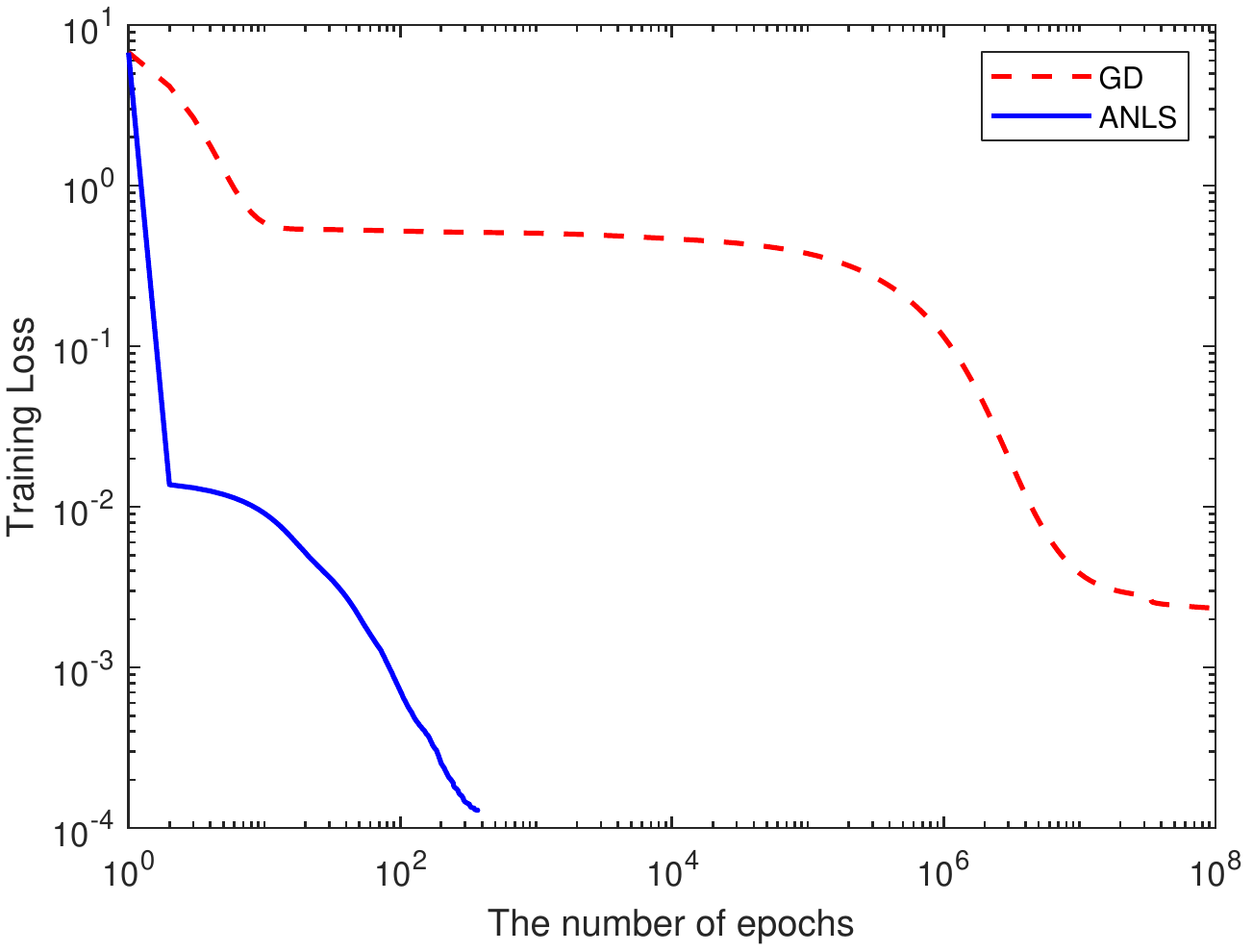}
		}
		\subfloat[ ]{\label{fig:moti2:b}
			\includegraphics[width=6.3cm, height=4.5cm]{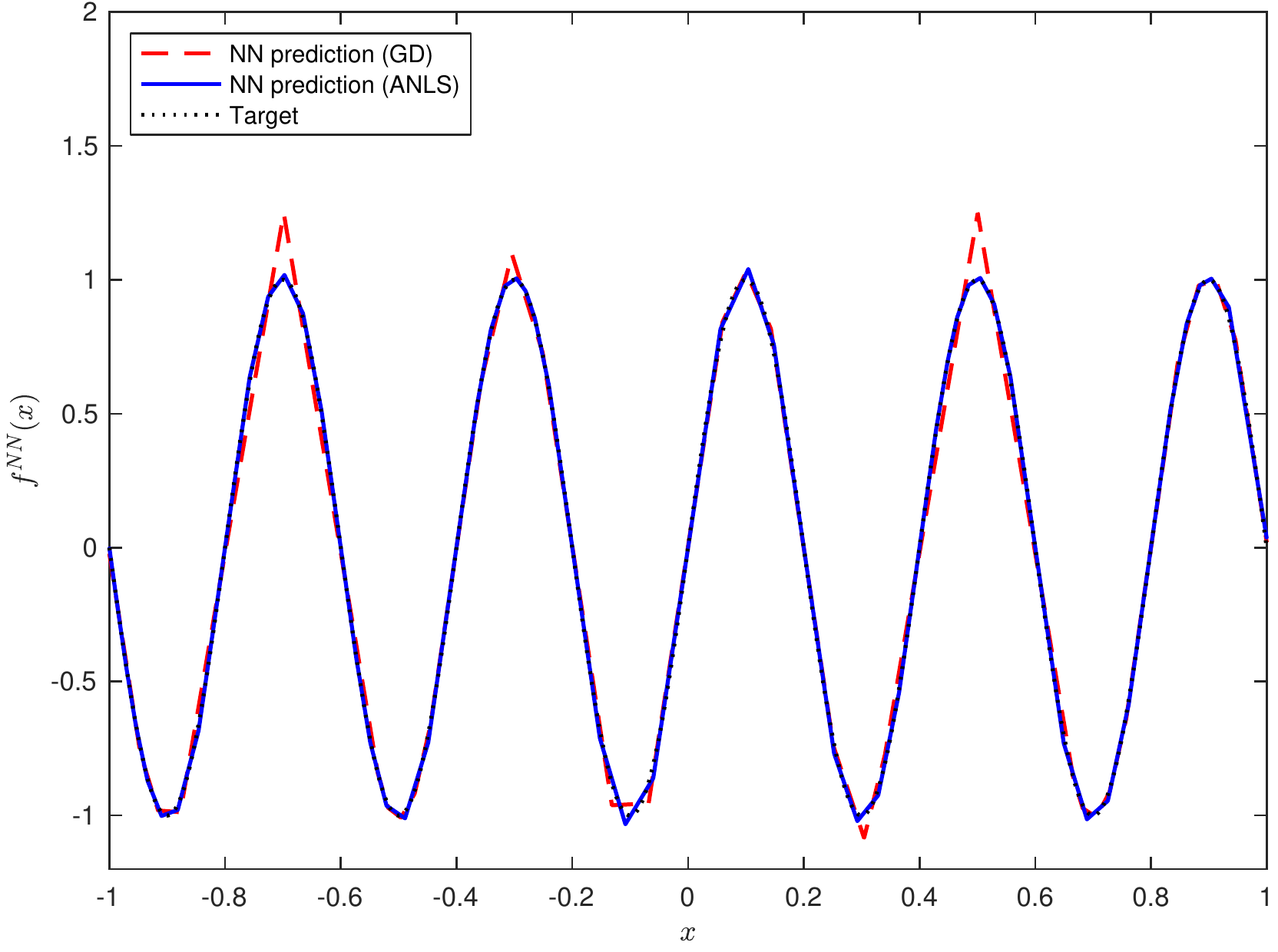}
		}
	}
	\caption{Left: Training loss obtained by gradient descent (GD) and the proposed method (ANLS) versus the number of epochs on the same learning task used in Figure~\ref{fig:motivation}.
	Right: The fully trained neural networks by both GD and ANLS.
	}
	\label{fig:motivation2}
\end{figure}

The results of applying ANLS to the 
same learning task used in Figure~\ref{fig:motivation}
are shown in Figure~\ref{fig:motivation2}.
Figure~\ref{fig:moti2:a} shows 
the loss versus the number of iterations compared with gradient descent,
while Figure~\ref{fig:moti2:b} shows the fully trained neural networks.

While the ANLS method utilizes a least squares solution step to determine the linear parameters $\bm{c}$, this is not the reason that it avoids plateau phenomenon.
Indeed, 
\cite{Cyr_19BoxInit} proposed a hybrid method that alternates between gradient descent and least squares, yet (as we will see) still exhibits plateau phenomena for essentially 
the same reasons as for pure gradient descent. 
The key feature of ANLS that sets it aside from alternative approaches
is the \emph{explicit adjustment of the activation pattern at each step},
which is designed to enable a quick exit from a plateau region.
An algorithm based on complete orthogonal decomposition \cite{Golub_Book12_Matrix}
is presented which enables one to efficiently identify the optimal adjustment to the activation patterns.

The rest of this paper is organized as follows. 
After describing the model problem in section~\ref{sec:setup},
we present the gradient flow analysis in section~\ref{sec:GDflow}.
The proposed active neuron least squares is presented in section~\ref{sec:method}.
Numerical examples are provided in section~\ref{sec:example} to demonstrate 
our theoretical findings and illustrate the effectiveness of the proposed method.

%

\section{Model Problem} \label{sec:setup}
Let us consider a univariate two-layer ReLU network
\begin{equation*}
f(x;\bm{\theta}) = \sum_{j=1}^n c_j\phi(w_jx-b_j), \qquad \bm{\theta} =  \{(b_j, c_j, w_j)\}_{j=1}^n,
\end{equation*}
where $\phi(x) = \max\{x,0\}$ is the rectified linear unit (ReLU) activation function.
Given a set of training data $\{(x_i,y_i)\}_{i=1}^m$, where $x_1 < x_2 < \cdots < x_m$,
we seek parameters $\bm{\theta}$ that minimize the loss function $\mathcal{L}$ defined by
\begin{equation} \label{def:Loss}
\mathcal{L}(\bm{\theta}) = \frac{1}{2}\sum_{i=1}^m (f(x_i;\bm{\theta})-y_i)^2,
\end{equation}
which measures the $\ell_2$ norm of the discrepancy between the output data and the prediction.
We shall see that many of the phenomena we wish to study
already manifest themselves in the special case where $w_j = 1$.
As a matter of fact, fixing $w_j=1$ does not change the expressivity of the neural networks, as shown by the following result:
\begin{proposition} \label{prop:w=1}
	Let $\Omega$ be a bounded interval and
	\begin{equation*}
	\begin{split}
	\mathcal{N} = \text{span}\{\phi(wx+b) | w, b \in \mathbb{R}, x \in \Omega \}, \qquad
	\mathcal{N}_{+} = \text{span}\{\phi(x+b) | b \in \mathbb{R}, x \in \Omega \}.
	\end{split}
	\end{equation*} 
	Then, $\mathcal{N} = \mathcal{N}_{+}$.
\end{proposition}
\begin{proof}
	Without loss of generality, let $\Omega=[0,1]$.
	Note that $\phi(wx+b) = \phi(b)$ if $w=0$
	and $\phi(wx+b) = |w|\phi(\text{sign}(w)(x+b/|w|))$ if $|w| \ne 0$.
	
	If $w=0$ and $b > 0$, observe that $\phi(b) = b = \phi(x + b) - \phi(x)$ for $x \in \Omega$.
	
	If $|w|\ne 0$, it suffices to show that any $\phi(-x+b)$ can be 
	expressed as a function in $\mathcal{N}_{+}$ on $\Omega$.
	Observing that 
	\begin{equation*} 
	\phi(-x+b) = \begin{cases}
	0, & \forall x \in \Omega, \text{if } b \le 0, \\
	\phi(x-b)-2\phi(x) + \phi(x+b), & \forall x \in \Omega, \text{if } b > 0
	\end{cases}
	\end{equation*}
	completes the proof.
\end{proof}

In view of Proposition~\ref{prop:w=1}, we may fix $w_j = 1$ for all $j$,
and consider networks of the form
\begin{equation} \label{def:NN}
f(x;\bm{b},\bm{c}) =\sum_{j=1}^n c_j\phi(x-b_j), \quad \text{where} \quad 
\bm{b} = \begin{bmatrix}
b_1 \\ \vdots \\ b_n
\end{bmatrix}, \quad
\bm{c} = \begin{bmatrix}
c_1 \\ \vdots \\ c_n 
\end{bmatrix}.
\end{equation}
The task of training the network is equivalent to selecting the positions of the nodes $\{b_j\}$ 
and the weights $\{c_j\}$
in a piecewise linear approximation \cite{He_etal_19ReLUFEM}. 
If the positions $\{b_j\}$ of the knots are specified, then the coefficients $\{c_j\}$ minimizing the loss can be obtained by solving a linear system of equations.
Here, however, we are also interested in determining the optimal knot placement which results in a highly non-linear minimisation problem for which we employ gradient descent to seek a minimum over both $\{b_j\}$ and $\{c_j\}$.

The network parameter $\bm{\theta} = \{(b_j,c_j)\}_{j=1}^n$ can be viewed as a vector in $\mathbb{R}^{2n}$.
Gradient descent proceeds iteratively starting with an initialization of the parameters $\bm{\theta}^{(0)}=\{(b_j^{(0)}, c_j^{(0)})\}_{j=1}^n$, and the $k$-th iteration updates the parameters according to the rule
$$
\bm{\theta}^{(k)} = \bm{\theta}^{(k-1)} - \eta_k \nabla_{\bm{\theta}} \mathcal{L}(\bm{\theta}^{(k-1)}), 
$$
where $\eta_k > 0$ is the learning rate at the $k$-th iteration
and $\nabla_{\bm{\theta}}$ denotes the gradient with respect to the free parameters $\bm{\theta}$.
In general, the learning rate should be chosen sufficiently small in order for the scheme to converge to a minimum.
In the limit $\eta_k \to 0$, we arrive at a continuous counterpart of the gradient descent scheme, in which the discrete iterates $\{\bm{\theta}^{(k)}\}$ are replaced by a function $\{\bm{\theta}(t), t \ge 0\}$
satisfying an initial value problem generated by the following system of first order autonomous ordinary differential equations (ODEs),
\begin{equation} \label{def:grad-flow}
\dot{\bm{\theta}}(t) = -\nabla_{\bm{\theta}} \mathcal{L}(\bm{\theta}(t)), \qquad 
\bm{\theta}(0) = \bm{\theta}^{(0)}.
\end{equation}
By the same token, the associated loss $\mathcal{L}(\bm{\theta}(t))$
may be regarded as a function of $t$ which, with a slight abuse of notation,
we shall henceforth denote as $\mathcal{L}(t)$.
Likewise, we shall speak of $t$ as a `time' although, strictly speaking, $t$ simply parameterizes the state of the system as the gradient descent updates proceed.
\section{Gradient Flow Analysis} \label{sec:GDflow}
The contribution to the error associated with the datum $(x_k,y_k)$ is measured by
$\text{loss}_{x_k}(t) = f(x_k;\bm{\theta}(t)) - y_k$,
and the corresponding residual vector is given by
\begin{align*}
\text{Loss}(t) = (\text{loss}_{x_1}(t), \cdots, \text{loss}_{x_m}(t))^T \in \mathbb{R}^m,
\end{align*}
so that $\mathcal{L}(t) = \frac{1}{2}\|\text{Loss}(t)\|_2^2$.
The equations governing the evolution of the knot $b_j$ and the coefficient $c_j$ in \eqref{def:grad-flow} are given by
\begin{equation} \label{def:bias-gradflow}
\begin{split}
\dot{b}_j(t) &= - \nabla_{b_j} \mathcal{L}(\bm{\theta}(t)) = c_j(t)\sum_{k=1}^m \text{loss}_{x_k}(t) \phi'(x_k - b_j(t)), \\
\dot{c}_j(t) &= - \nabla_{c_j} \mathcal{L}(\bm{\theta}(t)) = -\sum_{k=1}^m \text{loss}_{x_i}(t) \phi(x_k - b_j(t)), 
\end{split}
\end{equation}
subject to $b_j(0) = b_j^{(0)}$ and $c_j(0) = c_j^{(0)}$, $j=1,\cdots,n$. 
Here, $\phi'$ denotes the (left-continuous) Heaviside function
$\phi'(x) = \mathbb{I}_{x > 0}(x)$.

The discontinuity of the Heaviside function means that 
neither the Peano existence theorem nor the Picard–Lindel\"of theorem \cite{Teschl_12ODEs}
guarantee the existence of the solution to the system of ODEs \eqref{def:bias-gradflow}.
Instead, since the right-hand side of the ODEs \eqref{def:bias-gradflow} is measurable and locally essentially bounded,
it follows from \cite{Filippov_13differential,Bressan_98Discontinuous}
that for any initial point $\bm{\theta}^{(0)}$, 
there exists a Filippov solution of \eqref{def:bias-gradflow}
with initial condition $\bm{\theta}(0) = \bm{\theta}^{(0)}$.
A Filippov solution of \eqref{def:bias-gradflow}
is an absolutely continuous vector-valued function that satisfies the differential inclusion \cite{Aubin_12DiffInclusion} defined by the Filippov set-valued map.

Examining \eqref{def:bias-gradflow} shows a training point $x_k$ activates the neuron with bias $b_j$  if and only if $\phi(x_k - b_j)$ is non-zero, or equally well, $x_k > b_j$.
The set of all neurons activated by $x_k$ is given by 
\begin{equation*}
	A_k = \{ b_j | b_j < x_k \},
\end{equation*}
and the number of neurons activated by $x_k$ is $|A_k|$,
the cardinality of the set $A_k$.

\begin{figure}[htbp]
	\centerline{
		\includegraphics[width=12cm]{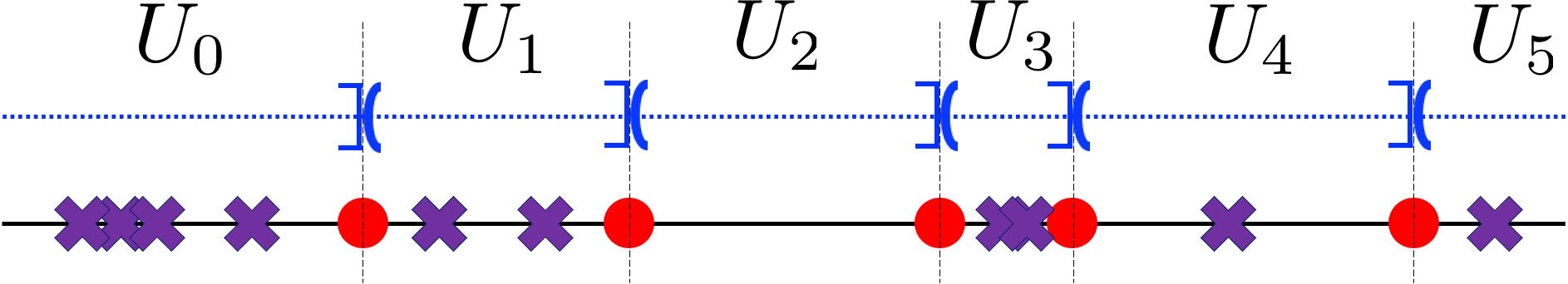}
	}
	\caption{Illustration of the set $\{U_l(t)\}_{l=0}^n$ for $n=5$.
	The red dots ($\textcolor{red}{\bullet}$) show the positions of the biases $\{b_j\}$. 
	The cross marks ($\textcolor[rgb]{0.4392, 0.1882, 0.6275}{\text{\ding{54}}}$) show the positions of the training points $\{x_k\}_{k=1}^{m}$ for $m=10$.
	The set $U_l(t)$ consists of the training points which lie 
	between the $l$-th and ($l+1$)-th largest biases.
	}
	\label{fig:set-Uk}
\end{figure}

The location of the $j$-th bias relative to the training points $\{x_k\}$
has a crucial impact on the flow.
For instance, in the trivial case where the initial values of the biases $\{b_j^{(0)}\}$ are chosen so that $\min_j b_j^{(0)} > \max_k x_k$, then the right hand of \eqref{def:bias-gradflow} vanishes identically meaning that 
$b_j(t) = b_j^{(0)}$ for all $t \ge 0$.
We seek to quantify this
by partitioning the training points into disjoint sets $U_l(t)$
at each time $t$, which consist of the training points $x_k$ that activate exactly
$l$ neurons, as follows:
\begin{equation} \label{def:U-set}
U_l(t) = \{x_k | |A_k| = l\}, \qquad l=0,1,\cdots,n.
\end{equation} 
The cardinality of $U_l(t)$ is denoted by $u_l(t) = |U_l(t)|$.
As a consequence, $\{U_l(t)\}_{l=0}^n$ is a partition of the training points $\{x_j\}_{j=1}^m$ and $\sum_{l=0}^n u_l(t) = m$.
In Figure~\ref{fig:set-Uk}, we illustrate $\{U_l(t)\}_{l=0}^n$ in the case $n=5$.

For each datum $(x,y)$, with slight abuse of notation,
we say $y \in U_l$ or $(x,y) \in U_l$ if $x \in U_l$.
The mean of the output data, and the mean and variance of the input data that correspond to the sets $\{U_l\}$ are defined respectively by
\begin{equation} \label{def:mean-variance-data}
\bar{y}_l = \frac{1}{u_l}\sum_{(x,y) \in U_l} y,
\qquad
\mu_{l} = \frac{1}{u_l}\sum_{x \in U_l} x,
\qquad
\sigma^2_{l} = \frac{1}{u_l}\sum_{x \in U_l} \left(x- \mu_{l}\right)^2.
\end{equation}

\subsection{Plateau Phenomenon}
Observe that the cardinalities $\{u_l(t)\}_{l=0}^n$ of the sets, being integer-valued, remain constant in time until
one or more of the biases $\{b_j\}$ crosses a data point,
wherein  the values of $\{u_l(t)\}$ jump to take new integer values.
The time periods, or stages, during which $\{u_l(t)\}$ remain constant play a crucial role in characterizing the evolution of the loss $\mathcal{L}(t)$ with time
as shown by the following result:

\begin{theorem} \label{thm:conti-loss-n}
	Suppose the biases and coefficients satisfy \eqref{def:bias-gradflow}, and let $[t_s,t_{s+1})$ be an interval where
	$\{u_l(t)\}$ remains constant.
	Let $\{\hat{e}^b_l, \hat{e}^c_l\}_{l=1}^n$ be vectors in $\mathbb{R}^m$
	defined as follows: for $j=1,\dots, m$,
	\begin{align*}
	[\hat{e}^b_l(t)]_j = \begin{cases}
	1 & \text{if } x_j \in U_l(t)  \\
	0 & \text{otherwise} 
	\end{cases},
	\qquad
	[\hat{e}^c_l(t)]_j 
	= \begin{cases}
	x_j - \mu_{l}  & \text{if } x_j \in U_l(t) \\
	0 & \text{otherwise} 
	\end{cases}.
	\end{align*}
	Let $V_t$ be the space spanned by $\{\hat{e}^b_l, \hat{e}^c_l\}_{l=1}^n$, $V_t^\perp$ be its orthogonal complement in $\mathbb{R}^m$,
	and $\Pi_{V}[x]$ be the orthogonal projection of $x$ onto $V$.
	Then, for $t \in [t_s,t_{s+1})$,
	\begin{align*}
	\left(\mathcal{L}(t_s) - Q_0(t_s)\right)e^{-\int_{t_s}^t \mathbf{r}_{\max}(v)dv}
	\le 
	\mathcal{L}(t) - Q_0(t_s)
	\le \left(\mathcal{L}(t_s) - Q_0(t_s)\right)e^{-\int_{t_s}^t \mathbf{r}_{\min}(v)dv}, 
	\end{align*}
	where $Q_0(t) = \frac{1}{2}\|\Pi_{V_{t}^\perp}[\text{Loss}(t)]\|^2$, and 
	$\mathbf{r}_{\min}(t)$ and $\mathbf{r}_{\max}(t)$
	are defined in Lemma~\ref{lem:def-M}.
	Furthermore, if $c_l(t) \ne 0$ for all $l$,
	$\mathbf{r}_{\min}(t)$ is strictly positive.
\end{theorem}
\begin{proof}
	The proof can be found in Appendix~\ref{app:thm:conti-loss-n}
\end{proof}


Theorem~\ref{thm:conti-loss-n} shows that for every time interval on which the activation patterns do not change, the component of the loss in the space $V_{t}$ decays exponentially at a rate determined by Lemma~\ref{lem:def-M}.
However, the component of the loss orthogonal to $V_{t}$ remains constant.
Throughout a time interval during which none of the biases cross a training point, the space $V_t$ does not change and the loss function decreases monotonically to the horizontal asymptote $Q_0(t_s)$.
This is precisely, the plateau phenomenon. 
The overall behavior of the loss $\mathcal{L}(t)$ consists of a number of stages during each of which $\mathcal{L}$ decreases towards a value $Q_0(t_s)$,
which depends only on the time $t_s$ at which the current stage was entered.
The value of the horizontal asymptote $Q_0(t_s)$ decreases monotonically with respect to $s$ 
(if the length of each time interval is sufficiently long).
Thus the graph of the loss with respect to the number of epochs 
takes the form of a decreasing staircase function, as shown in Figure~\ref{fig:motivation2} in section~\ref{sec:intro}.
The question of how long each interval should be 
in order for the horizontal asymptote $Q_0(t_s)$ to decrease monotonically
is addressed by the following result:

\begin{theorem} \label{thm:Q-decrease}
	Suppose the biases and coefficients satisfy \eqref{def:bias-gradflow}, and let $[t_s,t_{s+1})$ be an interval on which
	$\{u_l(t)\}$ remains constant.
	At time $t_{s+1}$, 
	suppose for some index $j$, 
	$u_l(t_s) = u_l(t_{s+1})$ for all $l$ but $j$ and $j+1$,
	and $|u_j(t_s) - u_{j}(t_{s+1})| = 1$.
	Suppose further that $t_{s+1}$ is large enough that 
	\begin{align*}
	e^{-\frac{1}{2}\int_{t_s}^{t_{s+1}} \bm{r}_{\min}(v)dv}
	< \frac{(1-|\zeta^T\psi|)|\langle \text{Loss}(t_s), \zeta \rangle|}{\|\Pi_{V_{t_s}} [\text{Loss}(t_s)] \|},
	\end{align*}
	where $\zeta \in V_{t_s}^\perp$ and $\psi \in V_{t_{s+1}}^\perp$ satisfying 
	\begin{equation*}
	\text{span}\{\zeta, \hat{e}^b_l(t_s), \hat{e}^c_l(t_s) : l=j, j+1 \}
	=\text{span}\{\psi, \hat{e}^b_l(t_{s+1}), \hat{e}^c_l(t_{s+1}) : l=j, j+1 \},
	\end{equation*}
	and $\hat{e}^b_l, \hat{e}^c_l$ are defined in Theorem~\ref{thm:conti-loss-n}.
	Then, $Q_0(t_s) > Q_0(t_{s+1})$.
\end{theorem}
\begin{proof}
	The proof can be found in Appendix~\ref{app:thm:Q-decrease}.
\end{proof}

\subsection{Convergence and Asymptotic Analysis}
\begin{figure}[htbp]
	\centerline{
		\includegraphics[width=4.3cm]{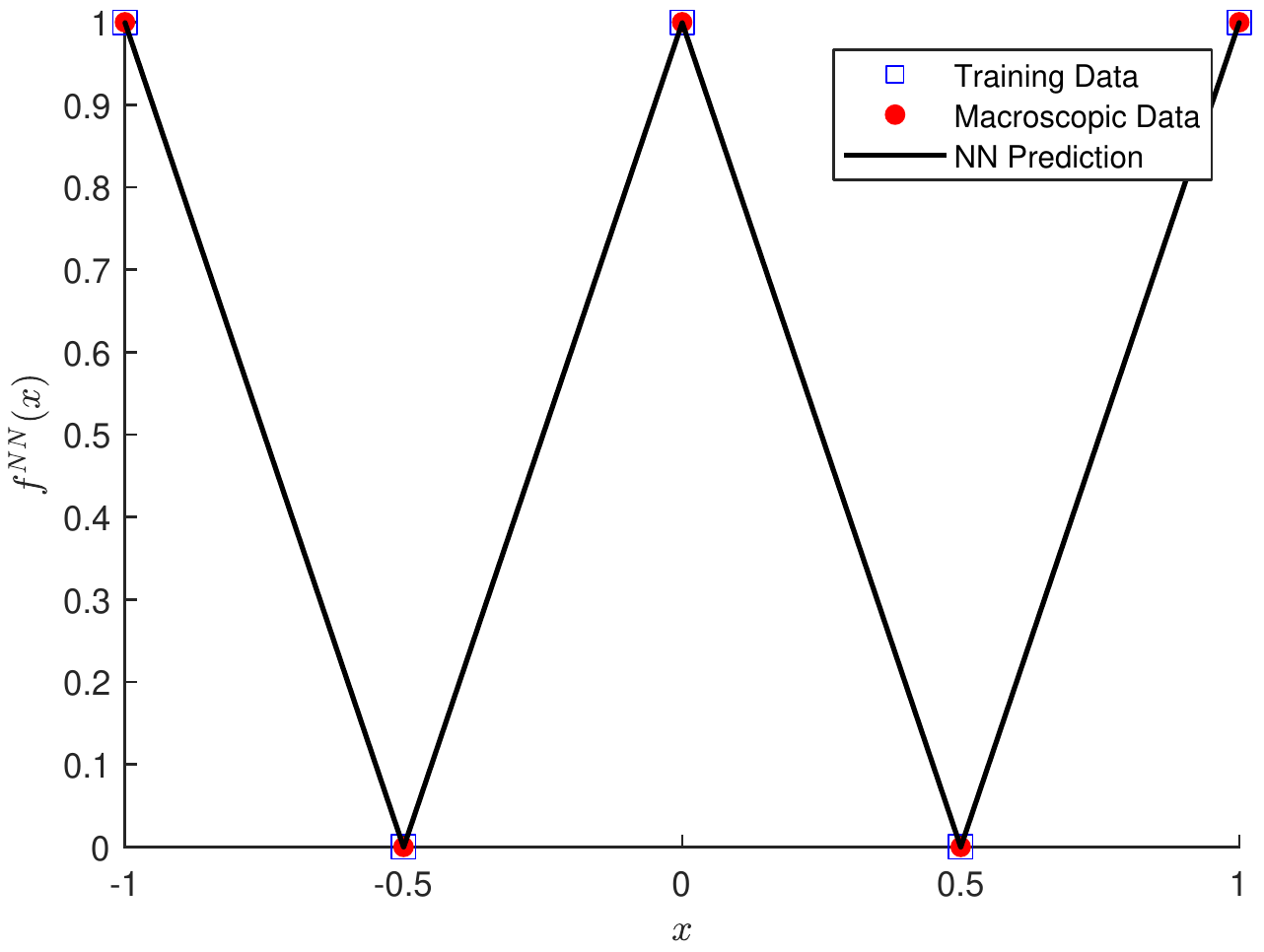}
		\includegraphics[width=4.3cm]{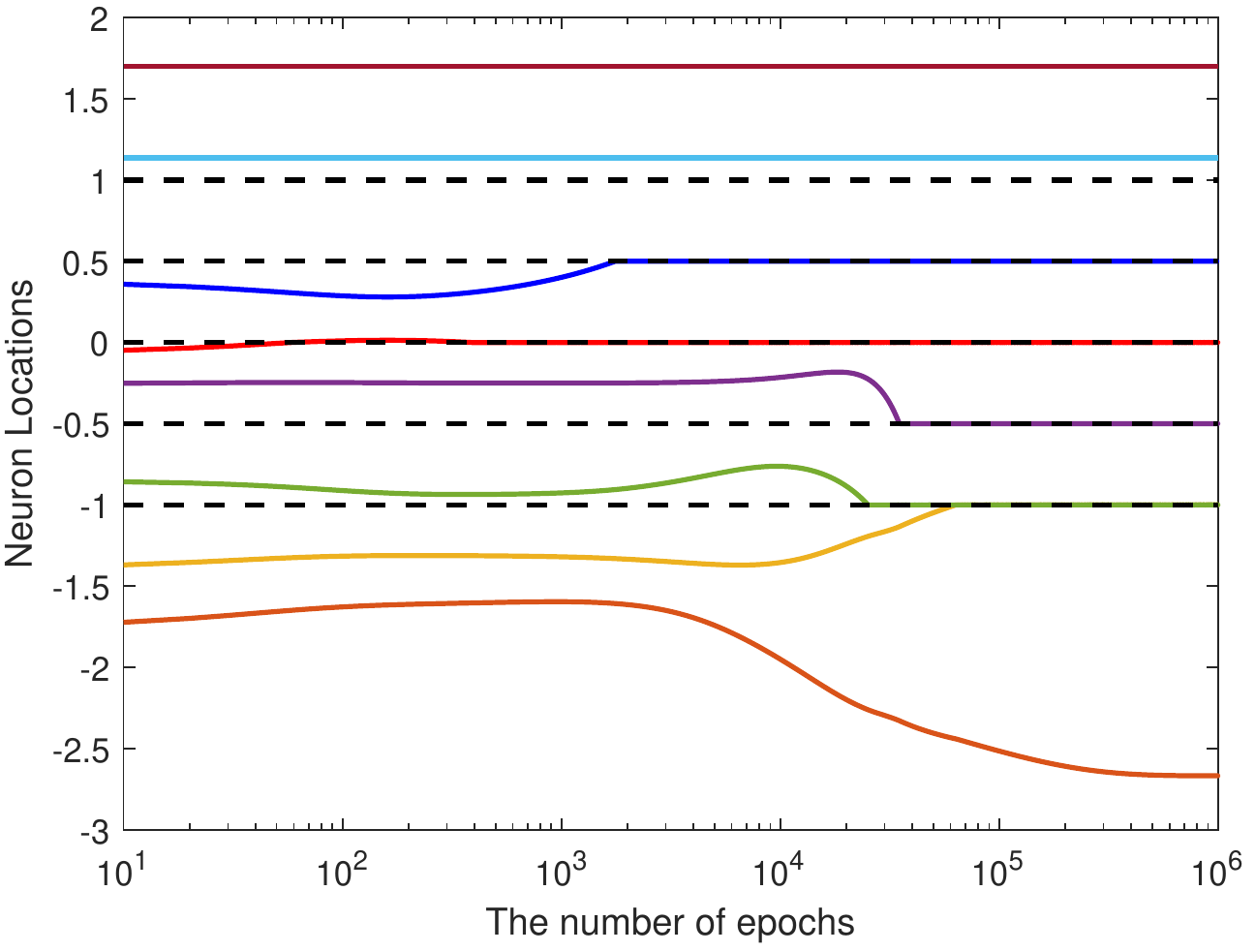}
		\includegraphics[width=4.3cm]{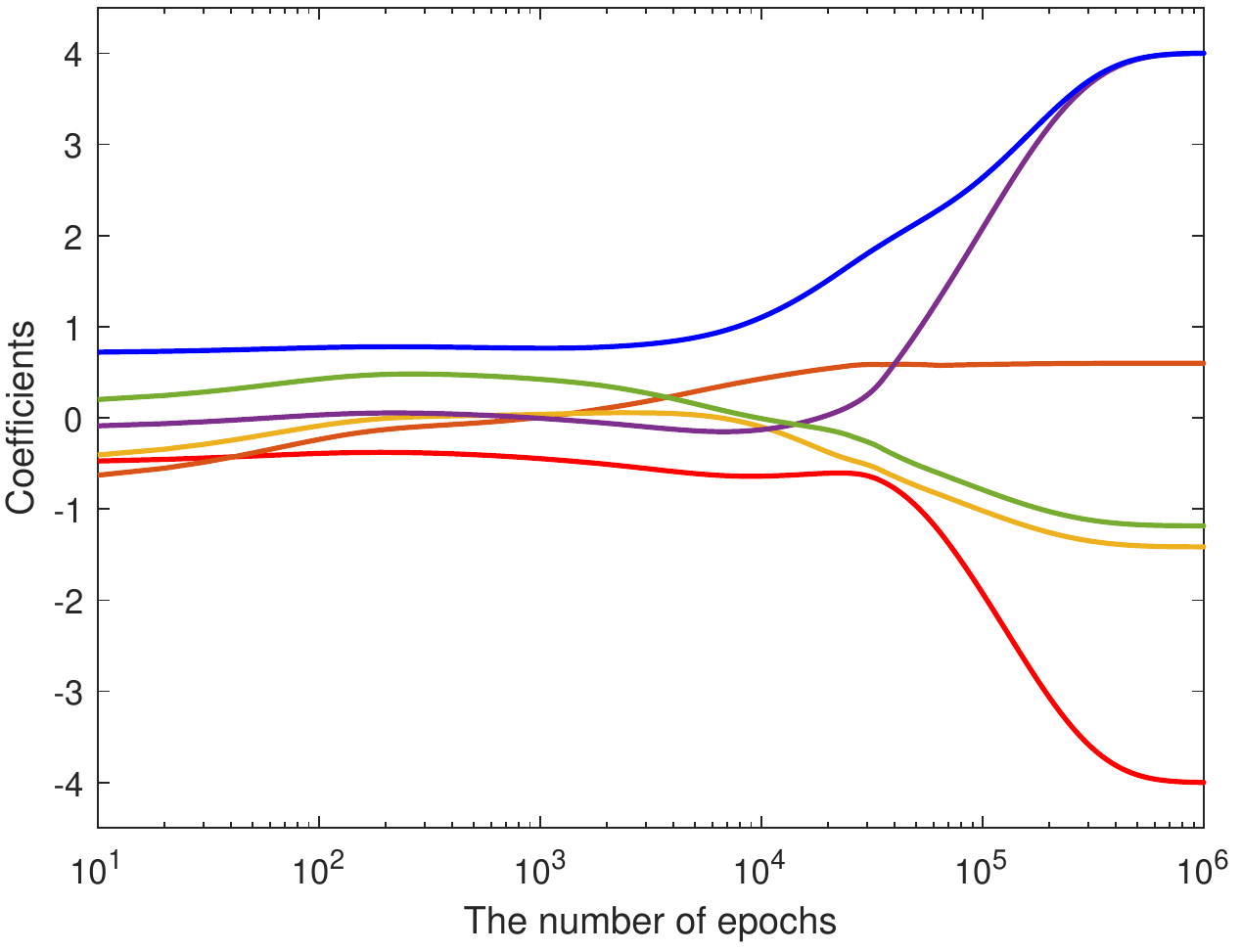}
	}
	\caption{Training results observed for fitting five training points
		with a two-layer ReLU network of width 8.
		The gradient decent is applied with a constant learning rate of $10^{-3}$.
		(Left) The final trained network.
		(Middle) The bias trajectories with respect to the number of epochs 
		along with the training input data (dashed lines).
		(Right) The coefficient trajectories with respect to the number of epochs.
	}
	\label{fig:motivation-graph}
\end{figure}

Next, we show the convergence of the weight and biases to a stationary (equilibrium) point of the gradient flow dynamics \eqref{def:bias-gradflow} under the condition that the sets \eqref{def:U-set} remain trapped in a given stage, i.e., $t_{s+1} = \infty$.
We note that it is quite possible (and common in practice) for the system to remain trapped in
a given stage for all $t \ge t_s$.
For example, in the trivial case mentioned earlier in which $\min_j b_j > \max_k x_k$
the system remains in the same stage for all $t$.
See also Figures~\ref{fig:motivation-graph} and ~\ref{fig:SinPI_W10N100-BC}.

\begin{theorem} \label{thm:convergence}
	Suppose the biases and coefficients satisfy \eqref{def:bias-gradflow}, 
	$\{u_l(t)\}$ is constant for $t \in [t_s, \infty)$,
	and $c_j(t)$ does not vanish at infinity for all $j$ with $u_j(t_s) \ne 0$. 
	Then, the biases and coefficients converge
	to a stationary point
	of 	the system \eqref{def:bias-gradflow}.
\end{theorem}
\begin{proof}
	The proof can be found in Appendix~\ref{app:thm:convergence}.
\end{proof}


As a first step towards characterizing fully trained networks, we identify necessary and sufficient conditions for the stationary (equilibrium) points of 
the system \eqref{def:bias-gradflow}.

\begin{lemma} \label{lemma:critical}
	$(\bm{b}^*,\bm{c}^*)$ is a stationary point of 
	the system \eqref{def:bias-gradflow}
	if and only if
	$(\bm{b}^*,\bm{c}^*)$
	satisfies
	\begin{equation} \label{cond:stationary}
	\begin{split}
	f(\mu_l;\bm{b}^*,\bm{c}^*) &= \bar{y}_{l}, \quad \forall l \text{ such that } u_{l} > 0, c^*_l \ne 0, \\
	\sum_{i=1}^{l} c_i^* &= 
	\frac{\frac{1}{u_l}\sum_{(x,y) \in U_l} xy - \mu_{l}\bar{y}_{l}}{\sigma^2_{l}},
	\qquad \forall l \text{ such that } u_l > 1,
	\end{split}
	\end{equation}
	where
	$\bar{y}_l, \mu_l, \sigma_l$ are defined in \eqref{def:mean-variance-data},
	and $f(x;\bm{b},\bm{c})$ is the ReLU neural network defined in \eqref{def:NN}.
	And the set of the stationary points of 
	the system \eqref{def:bias-gradflow} is nonempty.
\end{lemma}
\begin{proof}
	The proof can be found in Appendix~\ref{app:lemma:critical}.
\end{proof}

A linear stability analysis reveals that 
all eigenvalues of the linearized system 
at an equilibrium point are non-positive. 
The presence of eigenvalues with vanishing real part means that 
there is a non-trivial center manifold which we shall not 
investigate in detail in the current work. 
%

For the rest of the paper, we refer to the averaged data $\{(\mu_l, \bar{y}_l)\}_{l, u_l > 0}$ 
as the macroscopic data with respect to the sets $\{U_l\}$ \eqref{def:U-set}.
Also, we say a network is \emph{fully trained} if the parameter associated with it is a stationary point of the gradient flow dynamics \eqref{def:bias-gradflow}. 
An interesting feature of Lemma~\ref{lemma:critical} is that the equilibria depend only 
on averages of the training data defined in \eqref{cond:stationary}.
This means that if the training data were replaced by a smaller set which has the same average values defined in \eqref{cond:stationary}, then
the resulting equilibria are identical. 
This means that the network tends to fit an interpolant to this \emph{macroscopic} data
defined by \eqref{cond:stationary}.

Also, Lemma~\ref{lemma:critical} shows that 
the partial sums of coefficients should be some quantities that are  determined by the training data. 
Moreover, Lemma~\ref{lemma:critical}  provides a training stopping criterion that can be used for determining the termination of gradient descent.

In Figure~\ref{fig:motivation-graph},
we provide an illustration of Lemma~\ref{lemma:critical}
using a simple task of fitting five training data $\{(\frac{k-3}{2},\frac{1-(-1)^k}{2})\}_{k=1}^5$
using a two-layer ReLU network of width $8$.
On the left, the fully trained network is plotted along with both the original training data ($\square$) and the macroscopic data ($\circ$).
As expected by Lemma~\ref{lemma:critical}, we see that 
the fully trained network interpolates the macroscopic data, which is identical to the original training data in this case.
On the middle, the trajectories of biases are plotted with respect to the number of epochs.
For reference, the original five training data points are also plotted as dashed-lines.
We first see that the two neurons located beyond $1$ are dead and remain unchanged.
We observe that all neurons saturate as the number of epochs increases, which indicates that the sets $\{U_l\}$ remain trapped in a given stage. 
On the right, the trajectories of coefficients are plotted.
We see that non of them vanishes. 
Since the sets $\{U_l\}$ remain unchanged and all the coefficients are not vanishing as the number of epochs increases, as expected by Theorem~\ref{thm:convergence}, 
we observe the convergence of both biases and coefficients. 
To this end, we see that the fully trained network for this simple task is represented by the sum of five macroscopic neurons: 
$f^{\text{NN}}(x) = 4\phi(x-0.5)- 4\phi(x)+4\phi(x+0.5)+c_2\phi(x+1) +c_1\phi(x-b_1)$,
where $c_2\phi(x+1)$ is represented by the sum of two neurons (green, yellow),
and $c_1\phi(x-b_1)$ corresponds to the neuron colored orange.


The following theorem shows that the fully trained networks
can be described in terms of multiple least squares regression lines.

\begin{theorem} \label{thm:LSQ-Stationary}
	Let $(\bm{b}^*, \bm{c}^*)$ be a stationary point of the gradient flow dynamics \eqref{def:bias-gradflow}.
	Without loss of generality, let ${b}^*_l \le {b}^*_{l+1}$ for all $l$.
	For each $l$ with $u_l \ne 0$, 
	let $\{(x_s^l, \hat{y}^l_s)\}_{s=1}^{u_l}$ be a set of modified training data 
	where $(x_s^l, y_s^l) \in U_l$ and
	\begin{align*}
	\hat{y}_s^l = y_s^l - \sum_{j=1}^{l-1} c_j^*\phi(x_s^l - b_j^*),\quad \forall s= 1,\cdots, u_l.
	\end{align*}
	Then, for each $l$ with $u_l \ge 2$, $\ell_{LSQ}(x) = c^*_l(x -b^*_l)$ is the least squares regression line
	obtained by the data set $\{(x_s^l, \hat{y}^l_s)\}_{s=1}^{u_l}$.
	And for each $l$ with $u_l = 1$, there exists $\alpha_l \in \mathbb{R}$ such that 
	$c^*_l=	\alpha_l + \frac{x_1^l\hat{y}^l_1}{1+(x_1^l)^2}$ and
	$b^*_lc^*_l = \alpha_l x^l_1 -\frac{\hat{y}^l_1}{1+(x_1^l)^2}$.
\end{theorem}
\begin{proof}
	The proof can be found in Appendix~\ref{app:thm:LSQ-Stationary}.
\end{proof}

%

Theorem~\ref{thm:LSQ-Stationary} shows that 
a fully trained network consists of consecutive least squares regression lines, each of which is obtained by 
fitting the modified data from one of sets in $\{U_l\}$.
Also, the neurons whose corresponding $U_l$ is empty
connect least squares lines without incurring discontinuities.

\begin{figure}[htbp]
	\centerline{
		\includegraphics[width=8cm]{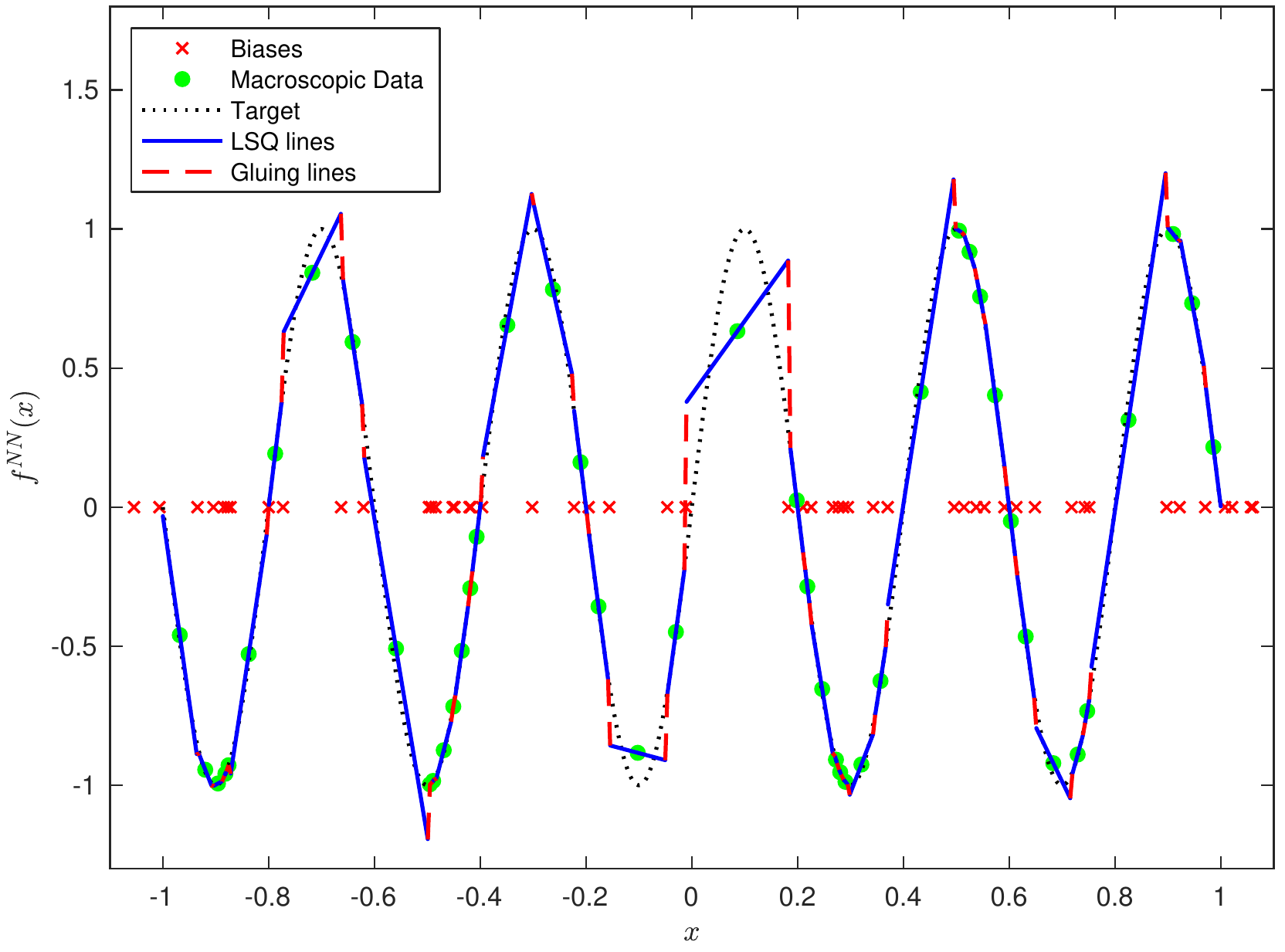}
	}
	\caption{A fully trained ReLU network by Proposition~\ref{thm:fullyNN}
		for approximating  $\sin(5\pi x)$.
		The sets $\{U_l\}$ are generated by the bias vector from $N(0,\sigma^2 I_n)$,
		where $\sigma = \sqrt{2}$, $I_n$ is the identity matrix of size $n$, and $n= 100$.
		The biases are marked as crosses.
		The least squares lines and the gluing lines are shown as solid and dashed lines, respectively. 
		The target function is shown as dotted line. 
	}
	\label{fig:FullyTrained}
\end{figure}

In conclusion, one may construct a fully trained neural network by gluing these least squares lines.
An explicit construction of such a network can be done by introducing extra neurons for
the purpose of gluing as described in Appendix~\ref{app:explicit-FTN}.
However, the construction may introduce seemingly artificial spikes. As a matter of fact, in Figure~\ref{fig:FullyTrained}, we illustrate the explicitly construction (Proposition~\ref{thm:fullyNN})
on the same learning task used in Figure~\ref{fig:motivation2}.
The sets $\{U_l\}$ are obtained from the initial bias vector, which is randomly generated from the normal distribution $N(0,\sigma^2 I_n)$, where $\sigma = \sqrt{2}$ and $n=100$.
We clearly see that the resulting network contains many spikes and does not approximate the target function very well. 
However, we emphasize that it is a fully trained network 
and no further improvement would be obtained by continuing the gradient descent.

%

\section{Avoidance of Plateau Phenomenon} \label{sec:method}
Based on the insights learned in the previous section,
we propose a new training method that avoids the plateau phenomenon.
The main idea is to incorporate a mechanism whereby the training algorithm
can exit a plateau region within a single iteration. 
The proposed method is iterative and gradient-free. 
At each iteration, the method systematically generates a set of candidate parameters and selects the one with the smallest loss. 


\subsection{Rank-Deficient Least Squares Problem}
Firstly, suppose that a bias vector is given,
leaving one with the task of minimizing the loss \eqref{def:Loss}
with respect to $\bm{c}$.
The corresponding coefficients are obtained by solving the least squares problem
\begin{equation*} 
	\min_{\bm{c}} \|\bm{A} \bm{c} - \bm{y}\|^2, \qquad [\bm{A}]_{ij} = \phi(x_i - b_j), \qquad \bm{y} = \begin{bmatrix}
	y_1 \\ \vdots \\ y_m
	\end{bmatrix}.
\end{equation*}
We are primarily interested in the adaptive region in which the amount of data exceeds the number of neurons. 
Furthermore, owing to the ReLU activation function $\phi$, the matrix $\bm{A}$ is often rank deficient. We are therefore faced with 
a rank deficient least squares problem which has infinitely many solutions.
We regularize by seeking the minimum norm solution:
\begin{equation} \label{def:LSQ}
	\bm{c}_{\text{LSQ}} \impliedby \min_{\bm{c} \in \mathcal{S}} \|\bm{c}\|, \qquad
	\mathcal{S} = \left\{\bm{c} \in \mathbb{R}^n : \|\bm{A}\bm{c} - \bm{y}\| = \min_{\tilde{\bm{c}}} \|\bm{A}\tilde{\bm{c}} - \bm{y}\|  \right\}.
\end{equation}
The solution to the problem is given by $\bm{c}_{\text{LSQ}} = \bm{A}^{\dagger}\bm{y}$, where $\bm{A}^{\dagger}$ is the Moore–Penrose inverse of $\bm{A}$.

If the solution satisfies the linear system exactly, 
a zero training loss is obtained and 
the resulting parameter $(\bm{b},\bm{c}_{\text{LSQ}})$ produces a fully trained network.

 
\subsection{Generating candidate pairs} \label{subsec:generating}
As shown in Theorem~\ref{thm:conti-loss-n}, each plateau 
corresponds to a period during which the activation patterns do not change.
As demonstrated in Figure~\ref{fig:motivation}, 
the main cause of the slow down of gradient descent training
is the time spent in traversing a plateau before entering a new one. 
In order to quickly exit such a plateau, 
the activation patterns \eqref{def:U-set} have to change.
To this end, given a bias vector $\bm{b}$,
we seek for a candidate bias, which produces different activation patterns. 
We make the following rules to generate candidate bias vectors.

Let $\bm{b} = [b_1,\cdots, b_n]^T$ be a given bias vector such that  $b_1 < \cdots < b_n$.
We say  $\bm{b}'=[b_1',\cdots,b_n']^T$ is a candidate bias vector with respect to $\bm{b}$ if the following rules are satisfied:
\begin{enumerate}
	\item  $b_j' = b_j$ for all $j$ but one. That is, we only change one neuron at a time. 
	\item 
	$b_1' < \cdots < b_n'$.
	That is, we do not move a neuron in a way that the bias ordering is violated. 
	\item 
	Let $\{U_l\}$ and $\{U_l'\}$ be the sets defined in \eqref{def:U-set} by $\bm{b}$ and $\bm{b}'$, respectively. 
	If the $k$-th neuron has changed, i.e., $b_k \ne b_k'$, 
	we have $|U_k| = |U_k'|  \pm 1$.
	\item If $U_l = \emptyset$ for all $l \ge k$, 
	then $b_l' = b_l$ for all $l \ge k$. That is,
	we do not move dead neurons.
	\item 
	If $U_l = \emptyset$ for all $l > k$ and $U_k \ne \emptyset$,
	then $U_k' \ne \emptyset$. 
	That is, we do not move a neuron in a way that the neuron is dead.
	\item 
	Suppose $b_k' \ne b_k$. Then, 
	$b_k'$ is one of three points:
	(i) the middle point of two data points,
	(ii) the middle point of a data point and $b_k$,
	or (iii) the smallest data point minus a fixed small number, say, $10^{-8}$.
\end{enumerate}
For each candidate bias vector, one can solve 
the least squares problem \eqref{def:LSQ} to find its corresponding optimal coefficient vector.
This generates a single candidate pair. 

\begin{figure}[htbp]
	\centerline{
		\includegraphics[width=13cm]{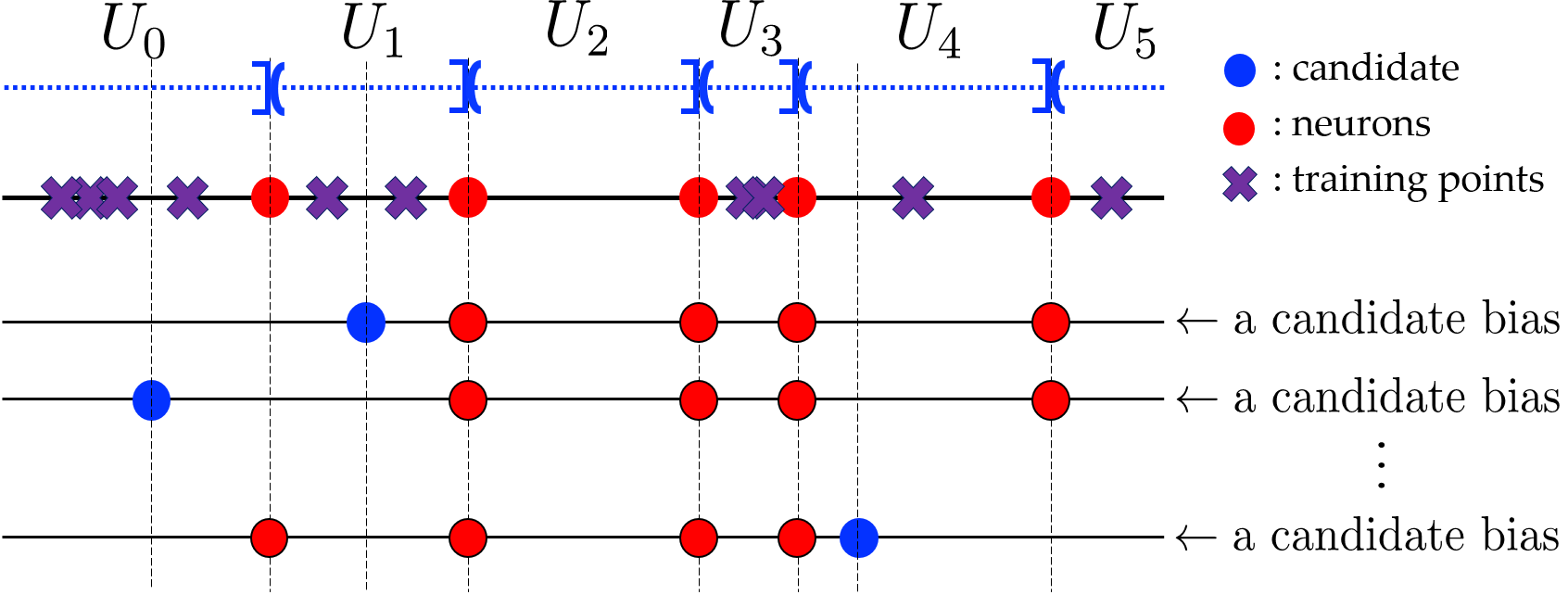}
	}
	\caption{Illustration of generating candidate biases by our rules for $n=5$.
		The red dots ($\textcolor{red}{\bullet}$) show the current positions of the biases $\{b_j\}$.
		The blue dots ($\textcolor{blue}{\bullet}$) show the feasible positions of the biases.
		The cross marks ($\textcolor[rgb]{0.4392, 0.1882, 0.6275}{\text{\ding{54}}}$) show the positions of the training points $\{x_k\}_{k=1}^{m}$ for $m=10$.
	}
	\label{fig:candidate-ex}
\end{figure}

We illustrate the generation of candidate bias vectors in Figure~\ref{fig:candidate-ex}
where there are 5 neurons and 10 data points.
It can be seen that there are two locations that the first neuron (far left) could be located
and each location generates a candidate bias vector, while fixing all the other neurons.
Since $U_2$ is empty, there is only one location that the second neuron could be located. 
Similarly, the third- and the fourth neurons generate one and two candidate vectors, respectively.    
Finally, since $U_5$ contains only one training data point, 
if the fifth neuron were located beyond the point in $U_5$,
it will not be activated by all the points and become dead. 
Thus, the fifth neuron can generate only one candidate vector by moving itself to the left.
By counting all the cases, we see that there are total of 7 candidate bias vectors generated by our rules.

\subsection{Active Neuron Least Squares} \label{subsec:ANLS}
Given $\bm{b}^{(k)}$, 
by solving the least squares problem \eqref{def:LSQ},
its corresponding optimal coefficient vector $\bm{c}_{\text{LSQ}}^{(k)}$ is obtained.
Let 
$\bm{\theta}^{(k)}_{\text{LSQ}} = (\bm{b}^{(k)}, \bm{c}_{\text{LSQ}}^{(k)})$. 
Let $\bm{b}^{(0)}$ be an initialization of the bias vector.
Starting at $k = 0$, the following steps are repeatedly conducted until converge:
\begin{itemize}
	\item Step 1: 
	By following the rules described in section~\ref{subsec:generating}, generate all the possible candidate pairs with respect to  $\bm{b}^{(k)}$.
	Find the best candidate pair $\bm{\theta}^{(k)}_{\text{BC}} = (\bm{b}_{\text{BC}}^{(k)}, \bm{c}_{\text{BC}}^{(k)})$ that results in the smallest loss.
	\item Step 2: 
	If $\mathcal{L}(\bm{\theta}^{(k)}_{\text{LSQ}}) \le \mathcal{L}(\bm{\theta}^{(k)}_{\text{BC}})$,
	terminate the process and return $\bm{\theta}^{(k)}_{\text{LSQ}}$.
	Else, set $\bm{b}^{(k+1)}:= \bm{b}_{\text{BC}}^{(k)}$
	and back to Step 1 with $k \to k+1$.
\end{itemize}

We call the proposed method as the \emph{active neuron least squares} (ANLS). 
By the construction of the method, the convergence of the loss by ANLS readily follows.
\begin{proposition} \label{prop:ANLSQ-convg}
	Let $\{\bm{\theta}^{(k)}_{\text{LSQ}} \}_{k\ge 1}$ be a sequence of parameters
	generated by the active neuron least squares.
	Then, 
	$
	\lim_{k \to \infty} \mathcal{L}(\bm{\theta}_{\text{LSQ}}^{(k)}) = L^*
	$
	for some $L^* \ge 0$.
\end{proposition}
\begin{proof}
	Suppose for all $k$, $\mathcal{L}(\bm{\theta}^{(k)}_{\text{BC}}) < \mathcal{L}(\bm{\theta}^{(k)}_{\text{LSQ}})$.
	Unless, the proof is done. 
	Since $\{\mathcal{L}(\bm{\theta}_{\text{LSQ}}^{(k)})\}_{k \ge 0}$ is a monotone decreasing sequence that is bounded below, 
	we have the convergence of the loss.
%
\end{proof}

\subsection{Efficient Implementation}
The step 1 of ANLS requires one to solve multiple least squares problems to determine the best pair. 
This could be computationally expensive, especially when the number of data is large. 
We thus present an efficient implementation of Step 1 via complete orthogonal decomposition (COD) \cite{Golub_Book12_Matrix}.

For a matrix $\bm{A}$, let us denote the $i$-th row and the $j$-th column of $\bm{A}$
by $\bm{A}_{i,:}$ and $\bm{A}_{:,j}$, respectively.
\begin{theorem} \label{thm:LSQ}
	Let $\bm{A}$ be a matrix of size $m\times n$ with  $m \ge n$.
	Let us consider a COD of $\bm{A}$:
	\begin{align*}
		\bm{Q}^T\bm{AZ} = \hat{\bm{T}}, \quad
		\text{where} \quad 
		\hat{\bm{T}} = \begin{bmatrix}
		\bm{T} & \bm{0} \\ \bm{0} & \bm{0}
		\end{bmatrix},
		\quad r = \text{rank}(\bm{A}), 
		\quad
		\bm{Z} = \begin{bmatrix}
		\hat{\bm{Z}}, & \tilde{\bm{Z}}
		\end{bmatrix}
	\end{align*}
	$\bm{Q}\in \mathbb{R}^{m\times m}$ and $\bm{Z}\in\mathbb{R}^{n\times n}$
	are orthogonal matrices,
	$\bm{T} \in \mathbb{R}^{r\times r}$ is a triangular matrix,
	and $\hat{\bm{Z}} \in \mathbb{R}^{n\times r}$.
	Let $\bm{V}_l$ be a matrix of size $m\times n$ 
	whose entries are all zeros except for the $l$-th column.
	Let $\bm{v}_l$ be the $l$-th column of $\bm{V}_l$.
	Let 
	\begin{align*}
	\bm{Q}^T\bm{v}_l &= \begin{bmatrix}
	\bm{q}_l^{(1)} \\  \bm{q}_l^{(2)}
	\end{bmatrix}, 
	\quad
	\bm{Q}^T\bm{y} = 
	\begin{bmatrix}
	\bm{p}^{(1)} \\ \bm{p}^{(2)} 
	\end{bmatrix},
	\qquad 
	\bm{q}_l^{(1)}, \bm{p}^{(1)} \in \mathbb{R}^{r},
	\end{align*}
	and $\tilde{\bm{A}} = \bm{A} + \bm{V}_l$.
	Then, the solution to the least squares problem \eqref{def:LSQ} for $(\tilde{\bm{A}}, \bm{y})$ is given by 
	$
		\tilde{\bm{c}}_{\text{LSQ}} = \bm{Z}\begin{bmatrix}
		\bm{x}_l^{(1)} \\ \bm{x}_l^{(2)}
		\end{bmatrix}
	$,
	and its corresponding loss is
	\begin{align*}
	\|\tilde{\bm{A}}\tilde{\bm{c}}_{\text{LSQ}} - \bm{y}\|^2 =
	\begin{cases}
	\|\bm{p}^{(2)}\|^2 - \|\bm{q}_l^{(2)}\|^2(d_l^*)^2, &\text{if } \|\tilde{\bm{Z}}_{l,:}\| > 0 \\
	\|\bm{p}^{(2)}\|^2  + \|\bm{p}^{(1)}\|^2  - \bm{s}_l^T\bm{x}_l^{(1)},  &\text{if } \|\tilde{\bm{Z}}_{l,:}\| = 0
	\end{cases},
	\end{align*}
	where $d^*_l = \begin{cases}
	\frac{\langle \bm{p}^{(2)}, \bm{q}_l^{(2)}\rangle }{\|\bm{q}_l^{(2)}\|^2}, &\text{if } \|\bm{q}_l^{(2)}\| > 0 \\
	\hat{\bm{Z}}_{l,:}\bm{x}_l^{(1)} &\text{if } \|\bm{q}_l^{(2)}\| = 0
	\end{cases}$, 
	$\bm{x}_l^{(1)} = 
	\begin{cases}
	\bm{T}^{-1}(\bm{p}^{(1)} - d_l^*\bm{q}_l^{(1)}), & \text{if } \|\tilde{\bm{Z}}_{l,:}\| > 0 \\
	\bm{M}_l^{\dagger}\bm{s}_l, & \text{if } \|\tilde{\bm{Z}}_{l,:}\| = 0
	\end{cases}$,
	\begin{align*}
	\bm{x}_l^{(2)} &= 
	\begin{cases}
	\frac{d_l^* - \hat{\bm{Z}}_{l,:}\bm{x}_l^{(1)}}{\|\tilde{\bm{Z}}_{l,:}\|^2}(\tilde{\bm{Z}}_{l,:})^T,& \text{if } \|\tilde{\bm{Z}}_{l,:}\| > 0 \\
	\bm{0}, & \text{if } \|\tilde{\bm{Z}}_{l,:}\| = 0
	\end{cases}, 
	\end{align*}
	$\tilde{\bm{T}}_l = \bm{T} + \bm{q}_l^{(1)}\hat{\bm{Z}}_{l,:}$,
	$\bm{M}_l = \tilde{\bm{T}}_l^T\tilde{\bm{T}}_l + \|\bm{q}_l^{(2)}\|^2\hat{\bm{Z}}_{l,:}^T\hat{\bm{Z}}_{l,:}$, 
	and 
	$\bm{s}_l = \bm{T}^T \bm{p}^{(1)} + (\bm{y}^T\bm{v}_l)\hat{\bm{Z}}_{l,:}^T$.
\end{theorem}
\begin{proof}
	The proof can be found in Appendix~\ref{app:thm:LSQ}.
\end{proof}

\textit{Remark 1:} 
Let $\bm{b}'$ be a candidate bias vector with respect to $\bm{b}^{(k)}$, which only differs by the $l$-th entry.
By applying Theorem~\ref{thm:LSQ} with $[\bm{v}_l]_j = \phi(x_j - b'_l) - \phi(x_j - b_l^{(k)})$,
one can obtain the candidate pair and its corresponding loss.

\textit{Remark 2:} 
Since $\|\bm{y}\|^2 = \|\bm{p}^{(1)}\|^2 + \|\bm{p}^{(1)}\|^2$
and $\|\bm{v}_l\|^2 = \|\bm{q}_l^{(1)}\|^2 + \|\bm{q}_l^{(2)}\|^2$,
the first $r$ columns of $\bm{Q}$ in Theorem~\ref{thm:LSQ} is necessary
for the method.

\textit{Remark 3:} 
In the case where $\|\tilde{\bm{Z}}_{l,:}\| = 0$, 
the method still requires to solve a $r\times r$ linear system of equations.
However, this is a much smaller problem compared to the original least squares \eqref{def:LSQ}.

\section{Numerical Examples} \label{sec:example}
In this section, we provide numerical examples to verify our theoretical findings
and demonstrate the performance of our proposed training method (ANLS). 

For the tests, we employ a two-layer ReLU network of the form \eqref{def:NN}.
The coefficients and the biases are initialized according to the `He' initialization \cite{he2015delving}.
That is, $c_j$'s are iid from $N(0,\frac{2}{n})$
and $b_j$'s are iid from $N(0,2)$,
where $n$ is the number of neurons.
The gradient descent (GD) is applied with a constant learning rate of $10^{-3}$. No mini-batch is employed.


\subsection{Verification of Gradient Flow Analysis}
\begin{figure}[htbp]
	\centerline{
		\subfloat[ ]{\label{fig:SinPI:a}
			\includegraphics[height=2.35cm]{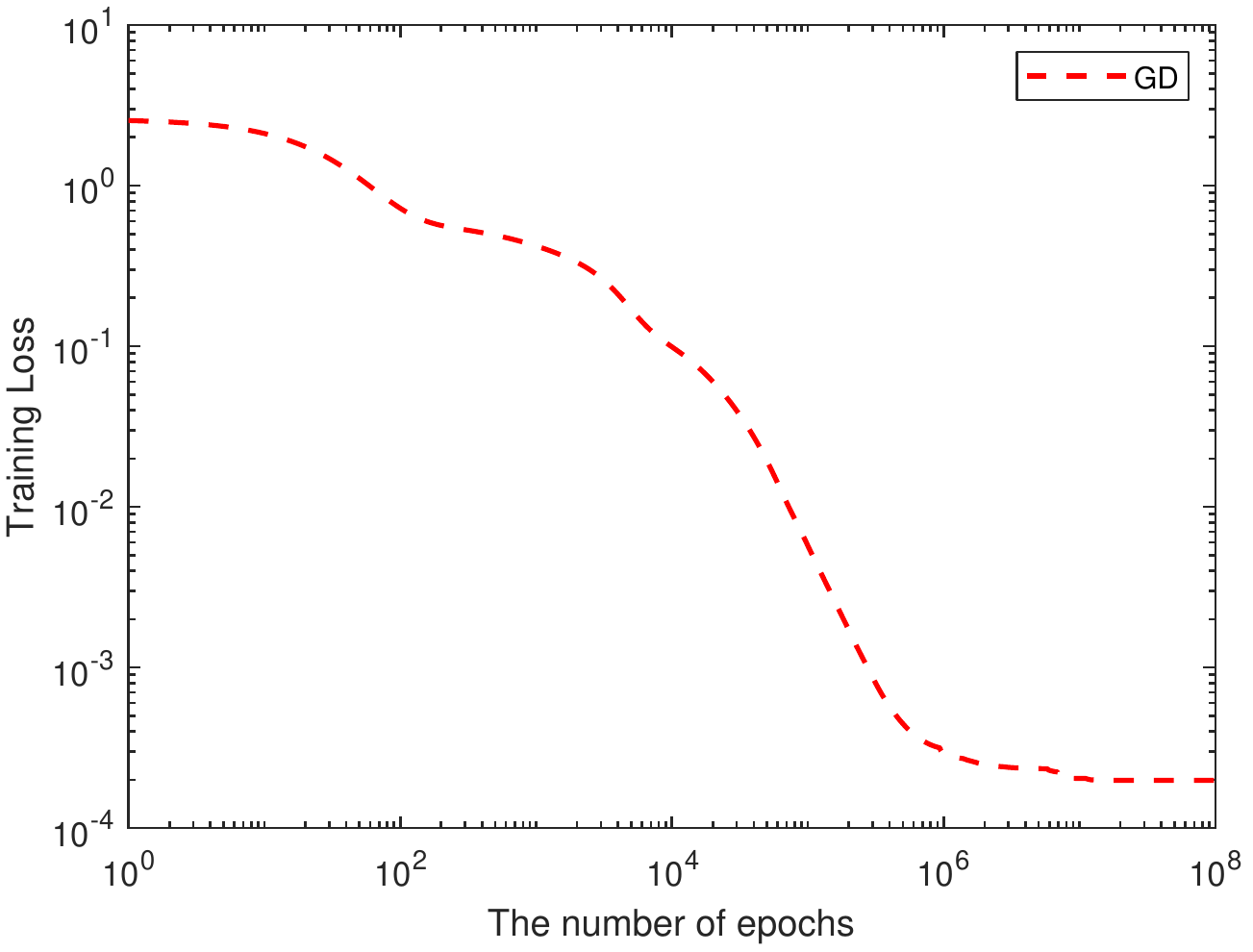}
		}
		\subfloat[ ]{\label{fig:SinPI:b}
			\includegraphics[height=2.35cm]{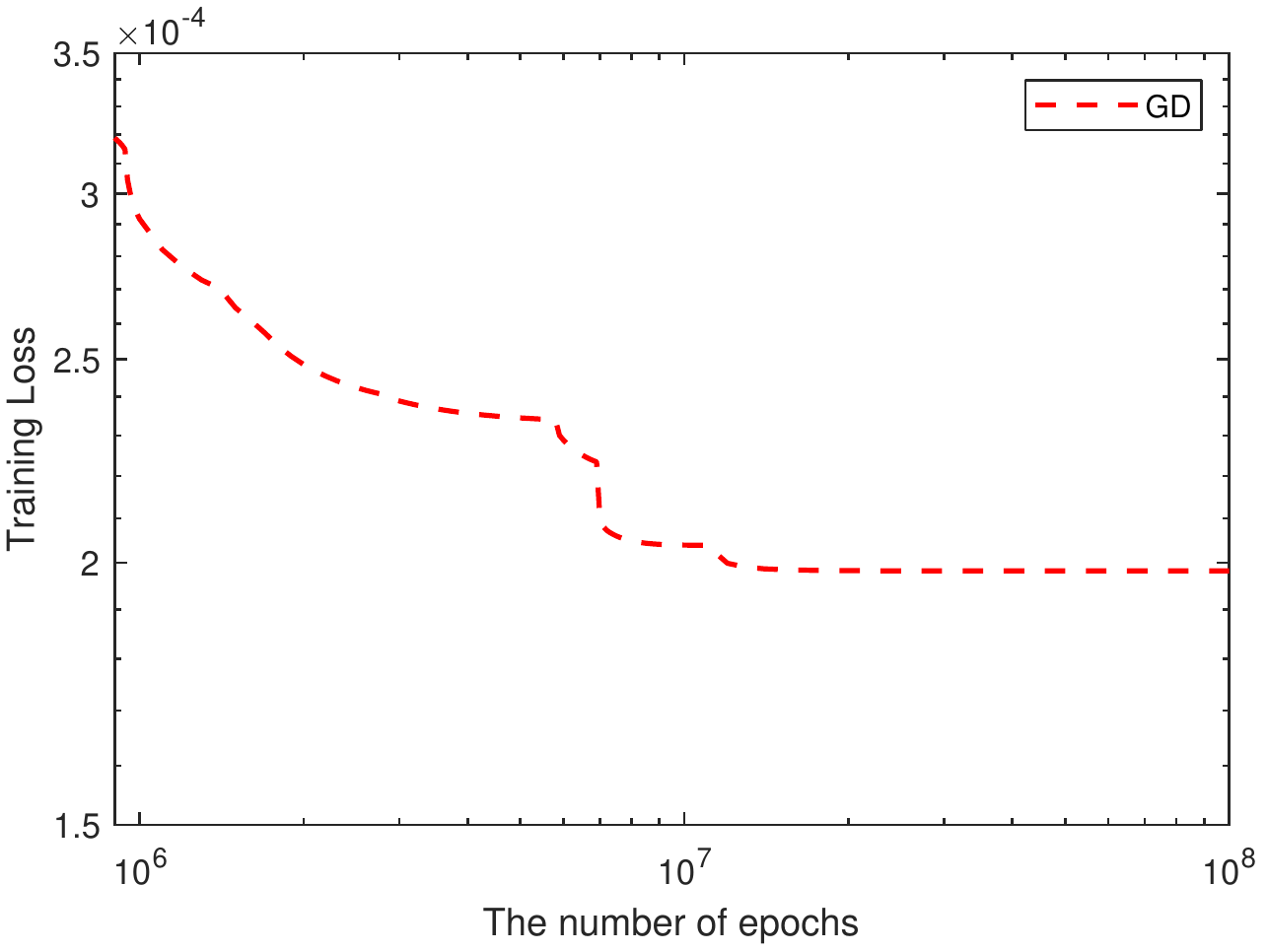}
		}	
		\subfloat[ ]{\label{fig:SinPI:c}
			\includegraphics[height=2.35cm]{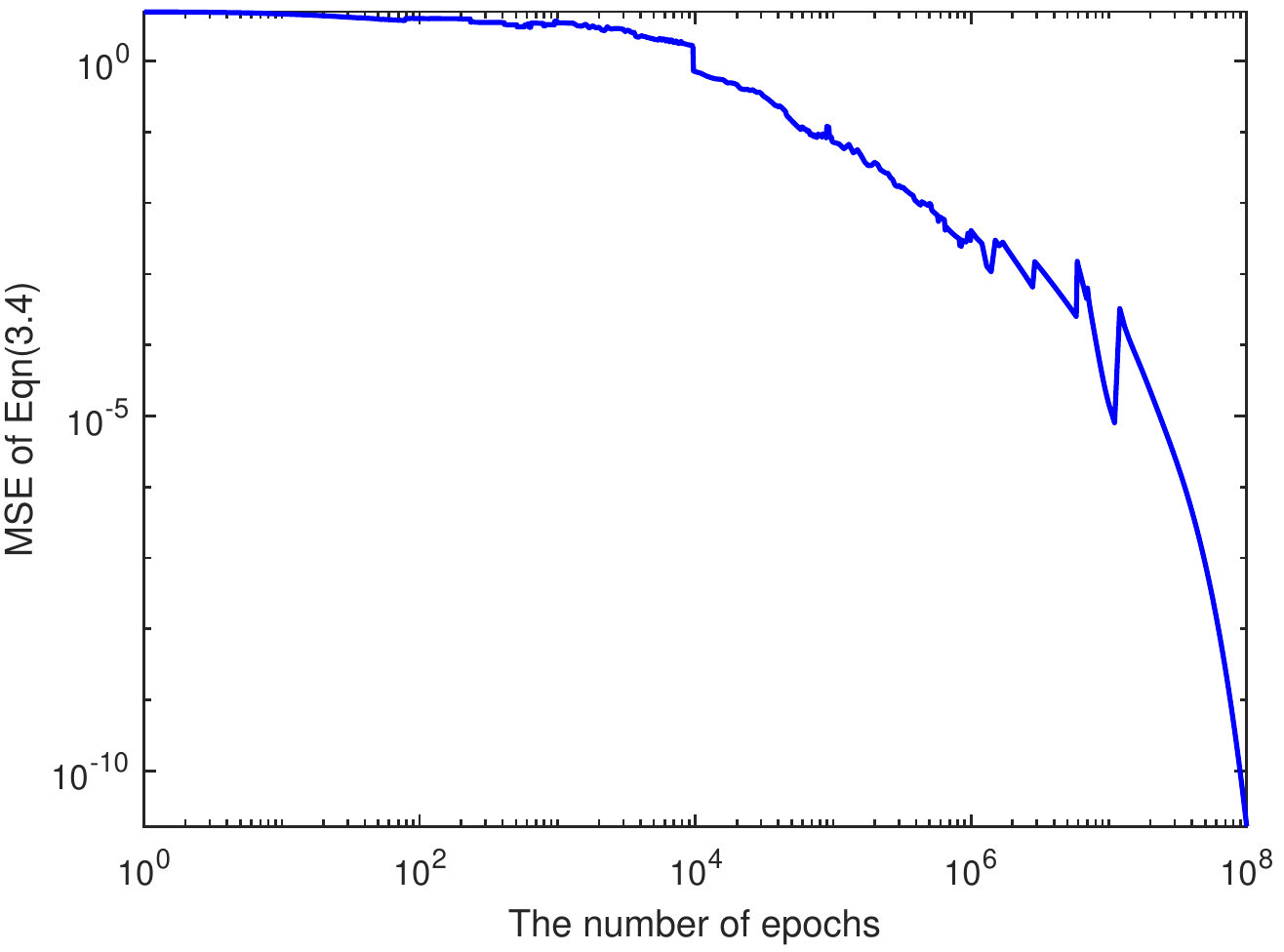}
		}
		\subfloat[ ]{\label{fig:SinPI:d}
			\includegraphics[height=2.35cm]{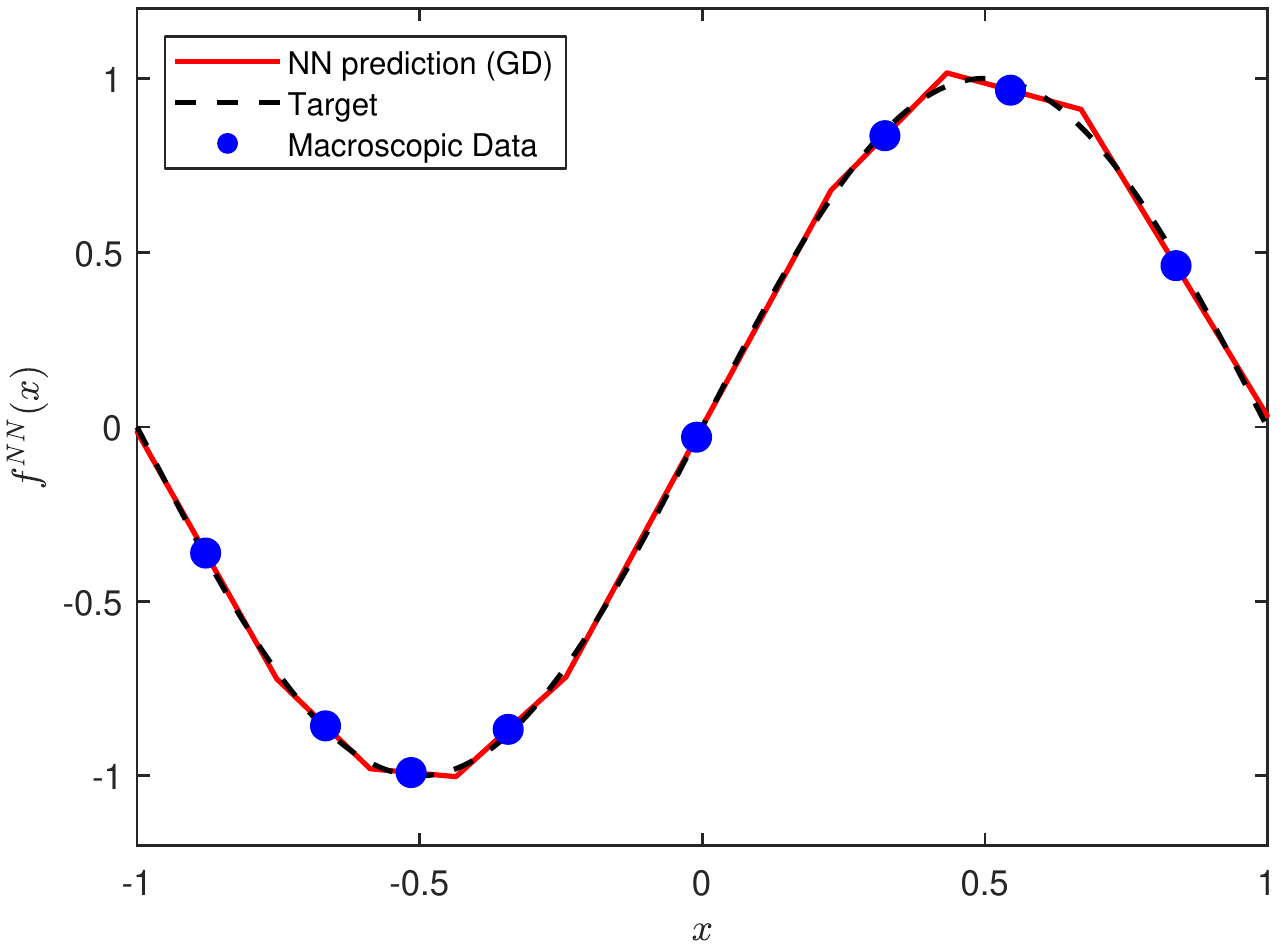}
		}
	}	
	\caption{Fig~\ref{fig:SinPI:a} and~\ref{fig:SinPI:b} show the training loss versus the number of GD iterations for approximating $\sin(\pi x)$
	using 100 equidistant data points from $[-1,1]$.
		The maximum number of GD iterations is $10^{8}$.
		A two-layer ReLU network of width 10 is employed.
		Fig~\ref{fig:SinPI:a} shows the entire loss trajectory.
		Fig~\ref{fig:SinPI:b} shows a partial loss trajectory where multiple plateaus are observed. 
		Fig~\ref{fig:SinPI:c} shows the mean squared error (MSE) of the system  \eqref{cond:stationary}.
		Fig~\ref{fig:SinPI:d} shows the fully trained network by GD.
		The macroscopic data are marked with filled circles.
	}
	\label{fig:SinPI_W10N100}
\end{figure}

First, we verify our gradient flow analysis (Lemma~\ref{lemma:critical} and Theorem~\ref{thm:convergence}).
We consider the problem of approximating a sine function:
\begin{equation} \label{test-func1}
\begin{split}
f_1(x) &= \sin(\pi x), \qquad x \in [-1,1].
\end{split}
\end{equation}
We use a ReLU network of width 10
and train it over 100 equidistant points from $[-1,1]$.
In Figure~\ref{fig:SinPI:a}, we show the training loss with respect to the number of epochs.
We clearly observe the plateau phenomenon.
GD training constitutes of multiple phases
and in each phase, we see that the loss decreases fast in the beginning and saturates later on.
The parts where the loss curve looks concave (convex) represent the fast (slow) decay of the loss.  
Figure~\ref{fig:SinPI:b} shows the later phase of training where multiple plateaus are observed. 
To quantify how far the parameter from being stationary, we report the mean squared error (MSE) of \eqref{cond:stationary} on Figure~\ref{fig:SinPI:c}.
We see that the MSE of  \eqref{cond:stationary} converges to 0, which indicates the completion of gradient descent training. In other words, the network is fully trained. 
However, even in this simple learning task, 
it takes more than 10 million iterations for GD to converge.
In Figure~\ref{fig:SinPI:d}, we plot the fully trained network.
As expected by Lemma~\ref{lemma:critical},
the fully trained network interpolates all the macroscopic data.

\begin{figure}[htbp]
	\centerline{
		\includegraphics[width=4.3cm]{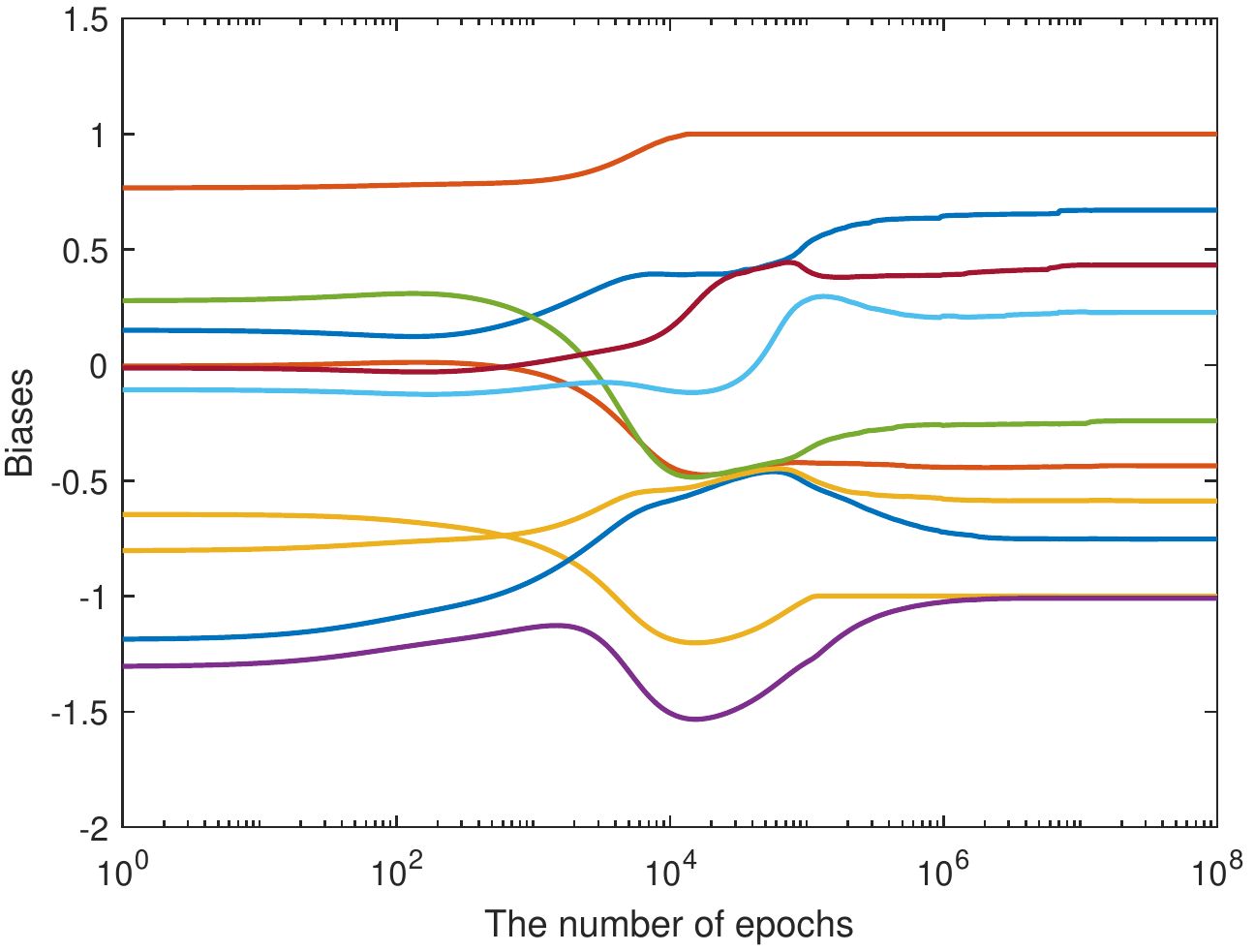}
		\includegraphics[width=4.3cm]{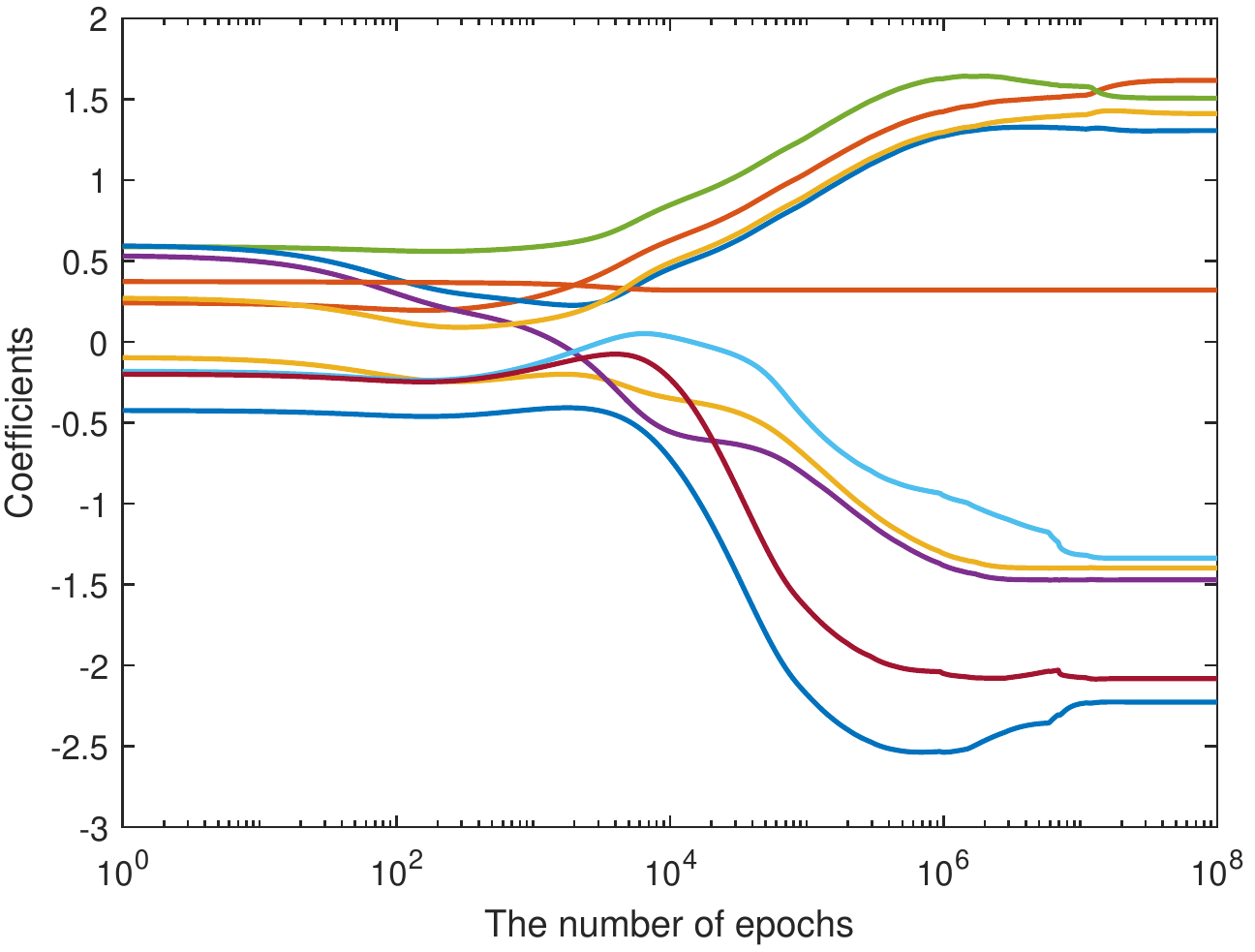}
		\includegraphics[width=4.3cm]{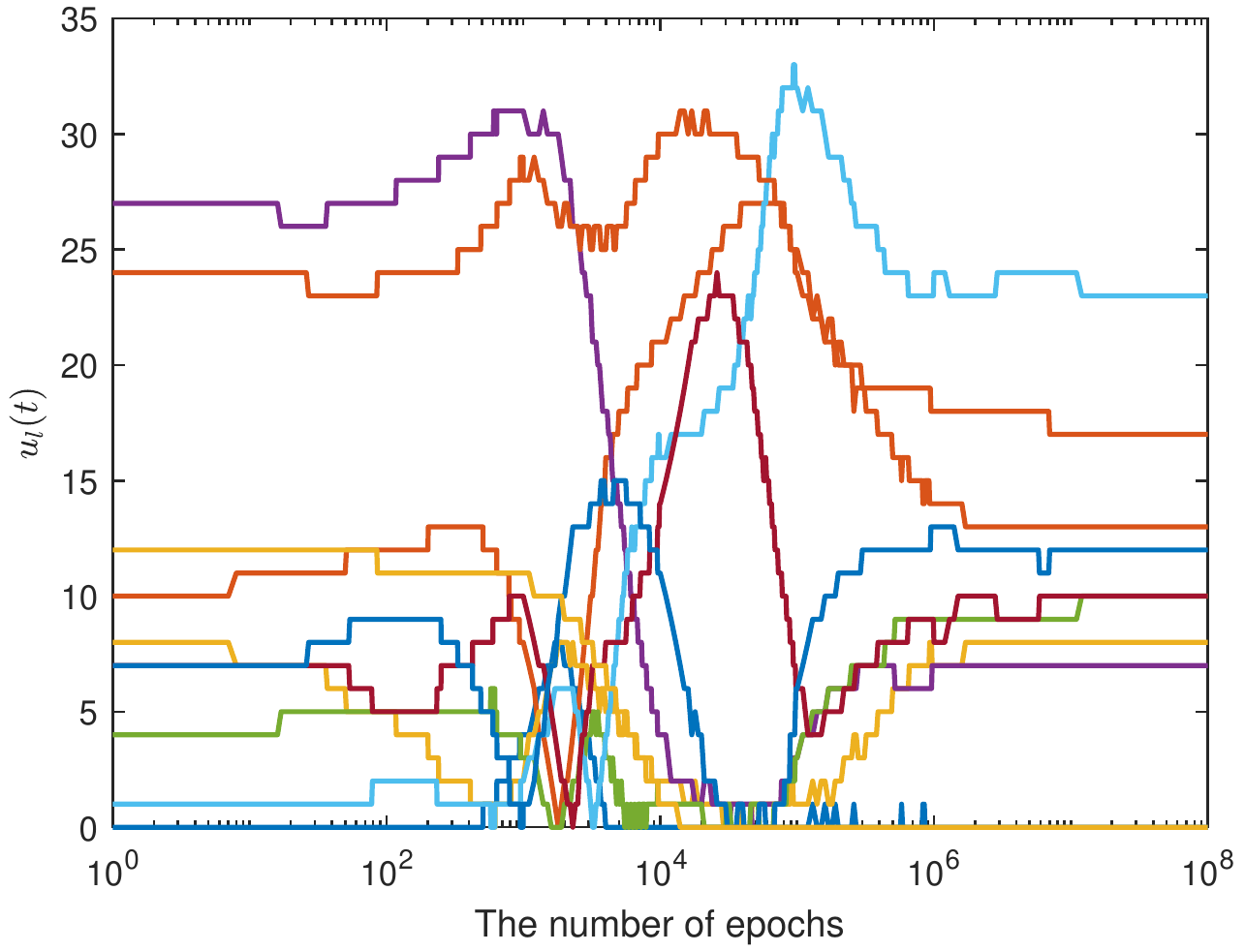}
	}	
	\caption{The trajectories of biases (left), coefficients (middle), and $\{u_l\}$ (right) with respect to the number of gradient descent iterations.
	}
	\label{fig:SinPI_W10N100-BC}
\end{figure}

Next, we demonstrate the convergence of the network parameters.
In Figure~\ref{fig:SinPI_W10N100-BC}, we show the bias and coefficient trajectories 
with respect to the number of epochs.
We clearly see the convergence of all the parameters.
In particular, we observe that the coefficients begin to converge
once the biases are stabilized. 
The stabilization of the biases implies
that the sets $\{U_l\}$ do not change anymore.
Indeed, on the right of Figure~\ref{fig:SinPI_W10N100-BC}, 
the trajectory of $\{u_l\}$ is plotted 
and we see that $\{u_l\}$ remains constant roughly after $2\times 10^7$ iterations.
Since non of coefficients are vanishing, 
by Theorem~\ref{thm:convergence}
the convergence of the parameters is guaranteed.

\subsection{Active Neuron Least Squares}
We demonstrate the performance of our proposed training method, the active neuron least squares (ANLS). 
For comparison, we report the results by the naive least squares (LSQ) \eqref{def:LSQ} 
where the bias vector is specified at the initialization
and the coefficient is obtained by solving its corresponding least squares problem
($\bm{\theta}_{\text{LSQ}}^{(0)}$ in section~\ref{subsec:ANLS}). 
Also, we report the results by \texttt{Adam} \cite{kingma2014adam}, one of the popular variants of gradient-descent,
with its default setup. 
Lastly, we report the results by the hybrid least squares/gradient descent \cite{Cyr_19BoxInit}
which updates the bias vectors via gradient descent (LS/GD) or \texttt{Adam} \cite{kingma2014adam} (LS/\texttt{Adam})
and then updates the coefficient by solving least squares.

\begin{figure}[htbp]
	\centerline{
		\includegraphics[width=6.5cm, height=5cm]{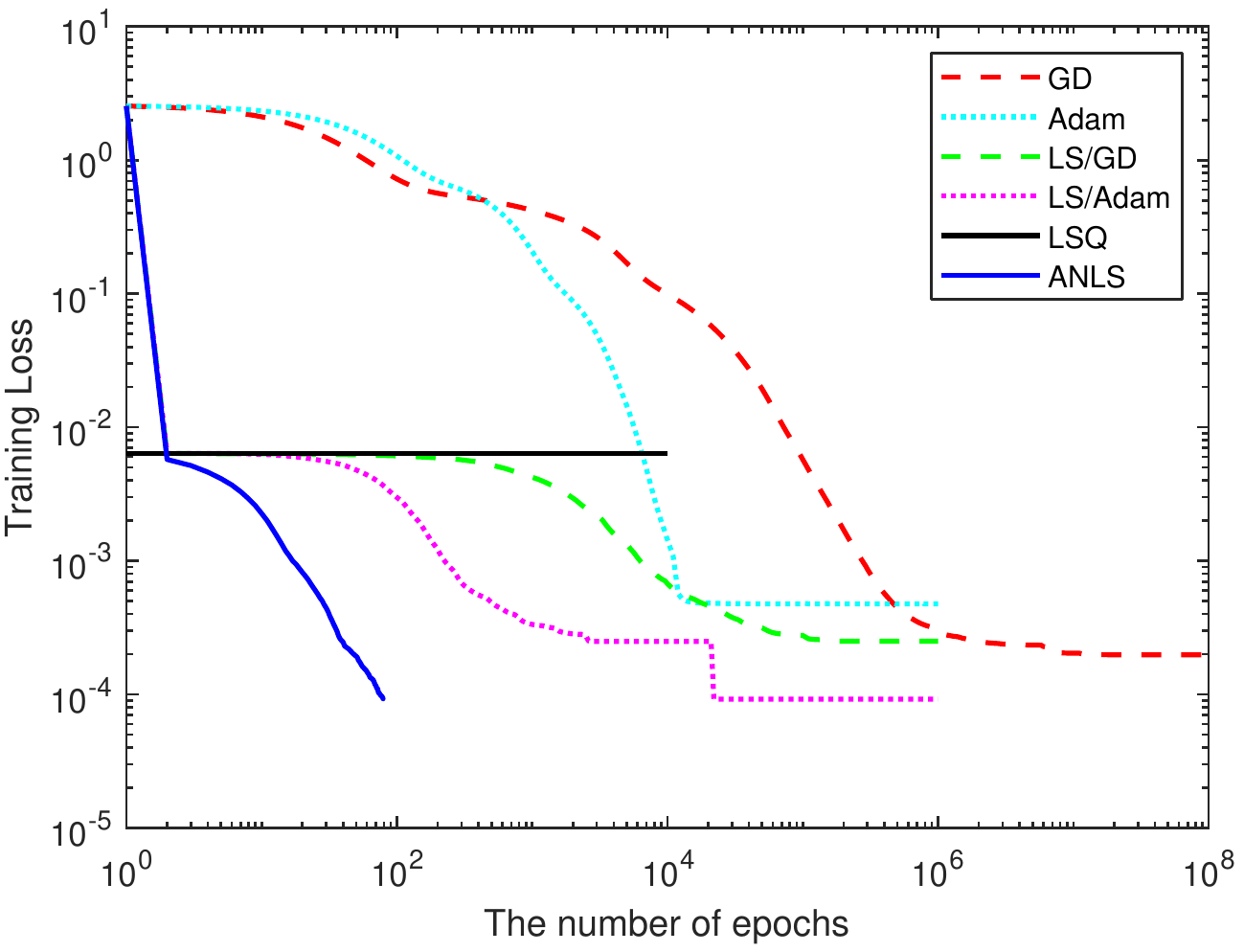}
		\includegraphics[width=6.5cm, height=5cm]{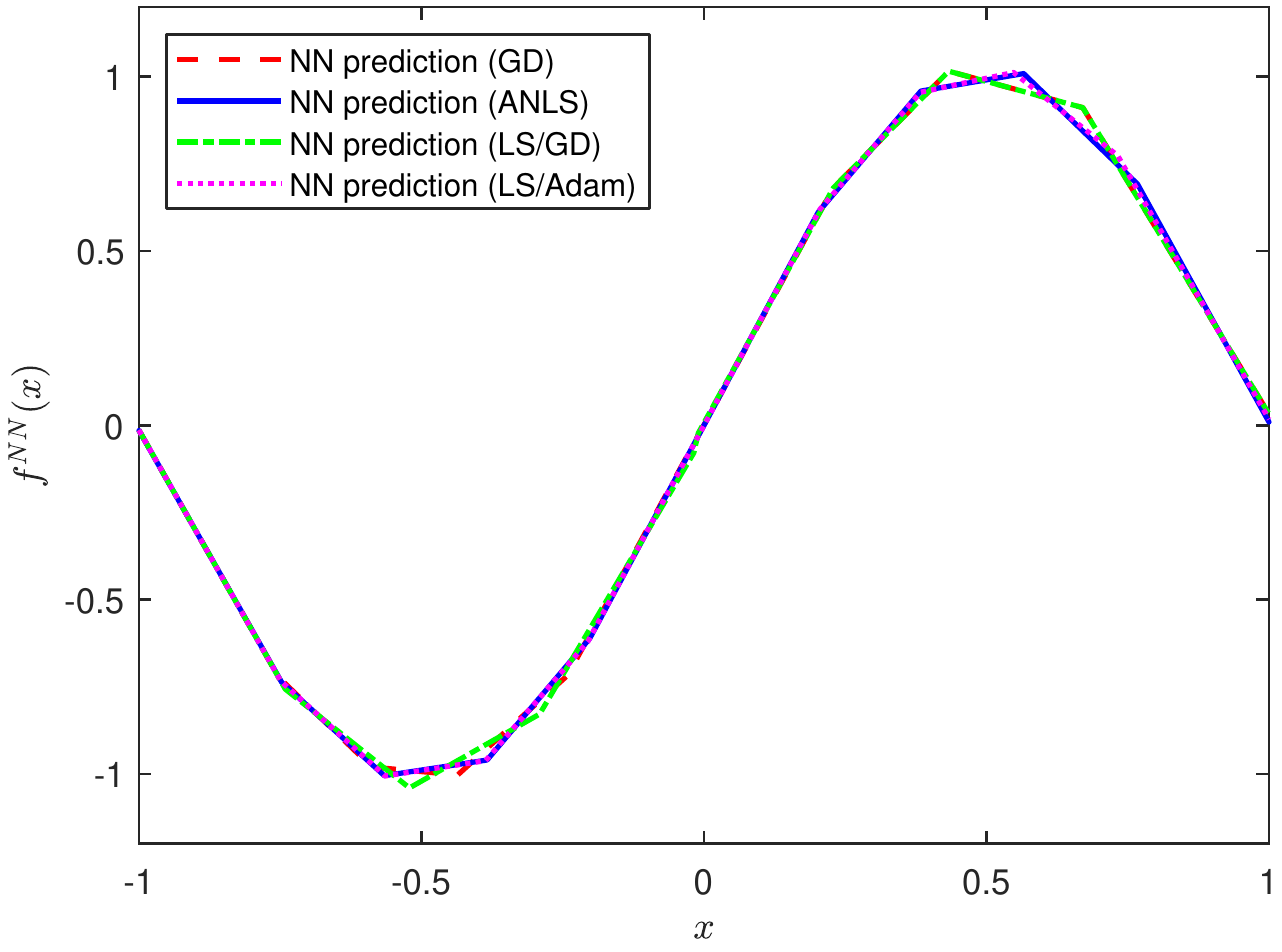}
	}	
	\caption{
		The results for approximating $f_1(x)$ \eqref{test-func1}
		with a two-layer ReLU network of width 10
		using 100 equidistant points from $[-1,1]$
		by ANLS, GD, \texttt{Adam}, LS/GD, LS/\texttt{Adam},
		and LSQ.
		Left: The loss with respect to the number of iterations.
		Right: The graphs of the fully trained networks.
	}
	\label{fig:SinPI_W10N100-ANLS}
\end{figure}

Firstly, let us consider the same learning task used in  Figure~\ref{fig:SinPI_W10N100}.
On the left of Figure~\ref{fig:SinPI_W10N100-ANLS},
the training loss by ANLS is plotted with respect to the number of iterations. 
We also report the results by GD, \texttt{Adam}, LS/GD, LS/\texttt{Adam} and LSQ. 
We see that ANLS achieves a much smaller loss value than those by LSQ.
Also, it is clearly observed that ANLS converges much faster than all the others in terms of the number of iterations. 
Within merely 100 iterations, ANLS already achieves the training loss of $10^{-4}$, however, all the others take more than at least 10,000 iterations
 to achieve a similar or larger loss value. 
And we observe that ANLS does not exhibit any stagnated stages in the training.
However all the gradient-based methods do exhibit multiple ones.
This indicates that ANLS indeed quickly exits plateaus. 
On the right, the fully trained network by ANLS is plotted. 
For reference, we also plot the fully trained network by GD, LS/GD, and LS/\texttt{Adam}.
We see that all the networks approximate the target function very well
including ANLS. 
This indicates the effectiveness of ANLS not only for fitting the training data but also for generalization.

\begin{figure}[htbp]
	\centerline{
		\includegraphics[width=8cm]{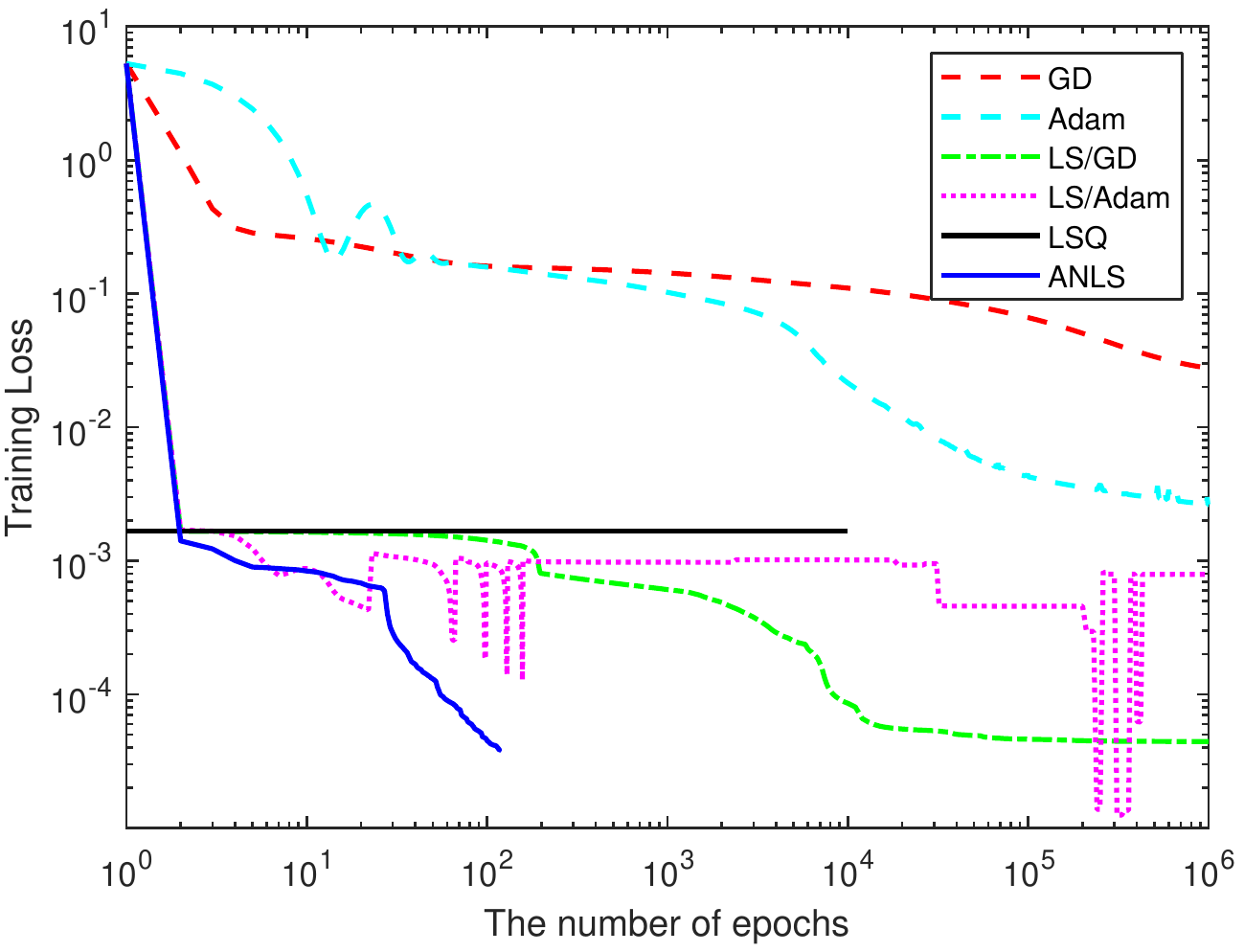}
	}
	\centerline{
		\includegraphics[width=6.5cm, height=5cm]{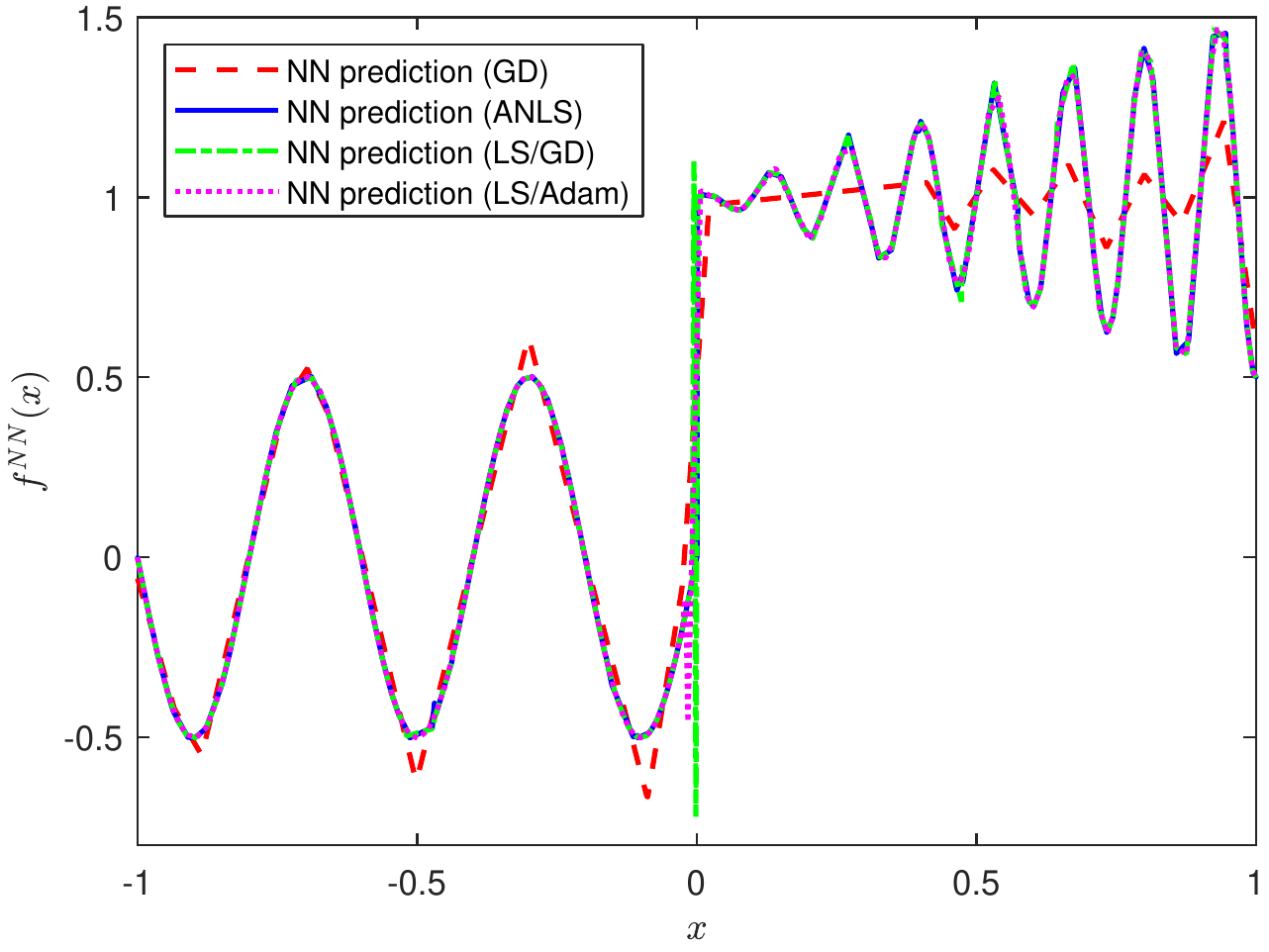}
		\includegraphics[width=6.5cm, height=5cm]{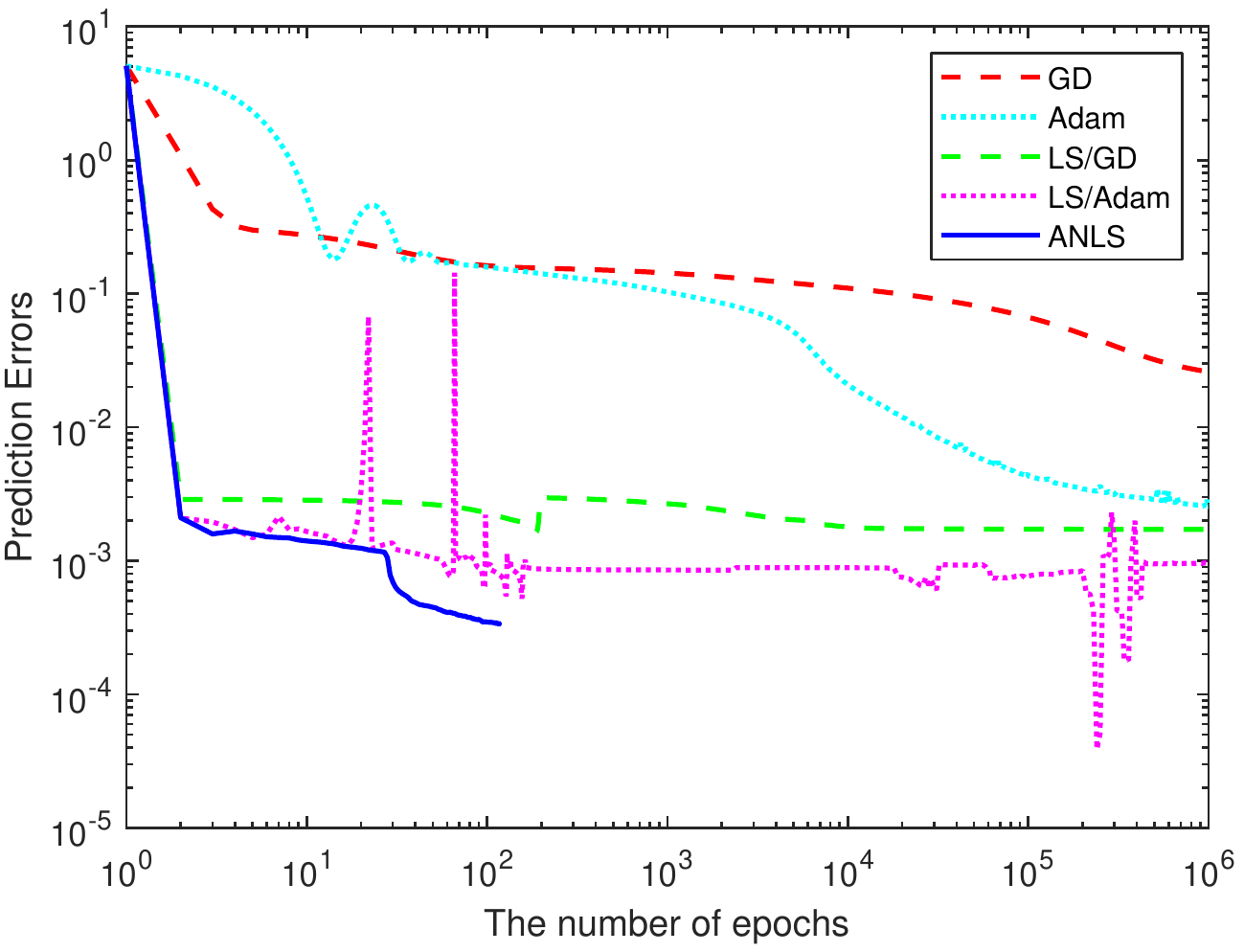}
	}	
	\caption{
		The results for approximating $f_2(x)$ \eqref{test-func2}
		with a two-layer ReLU network of width 300
		using 1,000 randomly uniformly drawn points from $[-1,1]$
		by ANLS, GD, \texttt{Adam}, LS/GD, LS/\texttt{Adam},
		and LSQ.
		Top: The loss versus the number of iterations.
		Bottom Left: The graphs of the fully trained networks.
		Bottom Right: The discrete $L_2$ prediction errors.
		The errors are computed on 10,000 equidistant points from $[-1,1]$.
	}
	\label{fig:TEST11}
\end{figure}

Next, we apply ANLS for approximating a discontinuous function:
\begin{equation} \label{test-func2}
\begin{split}
f_2(x) &= \begin{cases}
1 + 0.5x\cos(15\pi x), & \text{if } x > 0 \\
0.5\sin(5\pi x),  & \text{if } x \le 0
\end{cases}, 
\end{split}
\end{equation}
We employ a ReLU network of width 300
and train it over 1,000 randomly uniformly drawn training data points from $[-1,1]$.
On the top of Figure~\ref{fig:TEST11}, we show the training loss trajectories of ANLS and all the gradient-based methods.
Again, we clearly observe that ANLS converges much faster than all the comparisons.
The loss by ANLS converges approximately to $3\times 10^{-5}$ after around 100 iterations. 
On the other hands, the loss by GD is merely above $10^{-2}$
even after 5 million iterations,
the loss by \texttt{Adam} is roughly $2\times 10^{-3}$ after $1$ million iterations,
and the loss by the naive LSQ is approximately $10^{-3}$.
LS/GD achieves a similar loss level to ANLS at around 10,000 iterations.
And we observe multiple plateaus in the loss trajectories of GD, \texttt{Adam} and LS/GD.
However, ANLS does not show any long plateaus.
The loss trajectory of LS/\texttt{Adam} has many fluctuations and shows instability.
On the bottom left, the trained networks are plotted.
We see that the network trained by ANLS approximates the target function very well, whereas, those by GD, LS/GD, LS/\texttt{Adam} do not.
Especially, we observe an artificial spike around $x=0$ by both LS/GD and LS/\texttt{Adam}.
On the bottom right, the discrete $L_2$ prediction errors are reported with respect to the number of iterations. 
The errors are computed based on 10,000 equidistant points from $[-1,1]$.
We see that the prediction errors by GD, \texttt{Adam} and ANLS decrease
as the number of iterations grows and the smallest error is obtained by ANLS.
The error by LS/GD decreases in the early stage of training, however, 
it suddenly rapidly increases at around 200 iterations and slowly decreases again. 
 ANLS produces a network that has a smaller prediction error
than LS/GD, while both achieve a similar training loss value at the end of training.
This demonstrate a good generalization performance of ANLS.
Lastly, we observe that the trajectory of the prediction errors by LS/\texttt{Adam} is highly unstable and exhibit huge jumps.
%

\section{Conclusion} \label{sec:conclusion}
In this paper, we identified and quantified the root causes of plateau phenomenon
in gradient descent training of ReLU networks. 
We did not make any assumptions on the number of neurons relative to the number of training data.
By analyzing the gradient flow dynamics of univariate two-layer ReLU networks,
we showed that plateaux correspond to periods during which activation patterns remain constant, 
quantified convergence of the gradient flow dynamics,
and characterized stationary points in terms solutions of local least squares regression lines over subsets of the training data. 
Lastly, based on these conclusions, we proposed a new iterative training method, the active neuron least squares (ANLS) that is designed to quickly exit from a plateau
Numerical examples were provided to justify our theoretical findings and demonstrate the performance of ANLS. 

\appendix
\section{Proof of Theorem~\ref{thm:conti-loss-n}} \label{app:thm:conti-loss-n}

\begin{proof}
	It follows from the gradient flow dynamics \eqref{def:bias-gradflow}
	that  
	$
	\dot{\bm{b}}(t) = \bm{B}(t)\text{Loss}(t)$
	and $\dot{\bm{c}}(t) = -\bm{C}(t)\text{Loss}(t)$, 
	where 
	$\bm{B}(t)$ and $\bm{C}(t)$ are matrices of size $n \times m$ whose 
	$(j,k)$-components are defined to be
	$$
	[\bm{B}(t)]_{jk} = c_j(t)\phi'(x_k-b_j(t)), \qquad
	[\bm{C}(t)]_{jk} = \phi(x_k-b_j(t)),
	$$
	respectively, for $1\le k \le m,  1\le j \le n$.	
	Thus we have
	$$
	\frac{d}{dt}\text{Loss}(t) = -\bm{B}^T(t) \dot{\bm{b}}(t)
	+ \bm{C}^T(t) \dot{\bm{c}}(t)
	= -(\bm{B}^T(t)\bm{B}(t) + \bm{C}^T(t)\bm{C}(t))\text{Loss}(t).
	$$
	Hence,
	\begin{equation} \label{app:dt-loss}
	\begin{split}
	\frac{1}{2}\frac{d}{dt}\|\text{Loss}(t)\|^2 &= \langle \text{Loss}(t), \frac{d}{dt}\text{Loss}(t) \rangle = - \|\bm{B}(t)\text{Loss}(t)\|^2 - \|\bm{C}(t)\text{Loss}(t)\|^2.
	\end{split}
	\end{equation}
	
	Without loss of generality, 
	let us assume that $t \in [t_s, t_{s+1})$ on which 
	$\{u_l(t)\}_{l=0}^n$ is constant.
	Also, we assume that the neurons are ordered so that 
	$b_1(t) \le b_2(t) \le \cdots \le b_n(t)$.
	For simplicity, we also drop the expression of $t$ in $b_j, c_j$.
	We observe that
	\begin{align*}
	\bm{B}^T = \begin{bmatrix}
	c_{1}\xi_1^b & c_{2}\xi_2^b & \cdots & c_{n}\xi_n^b
	\end{bmatrix},
	\qquad
	\bm{C}^T= \begin{bmatrix}
	\xi_1^c & \xi_2^c & \cdots & \xi_n^c
	\end{bmatrix},
	\end{align*} 
	where  $\xi_k^b, \xi_k^c$ are vectors in $\mathbb{R}^m$ whose components are defined by
	\begin{align*}
	[\xi^b_k]_j = \begin{cases}
	1 & \text{if } x_j > b_{k} \\
	0 & \text{elsewhere}
	\end{cases},
	\qquad
	[\xi^c_k]_j =  \begin{cases}
	x_j - b_{k} & \text{if } x_j > b_{k} \\
	0 & \text{elsewhere}
	\end{cases}.
	\end{align*}
	\begin{lemma} \label{app:lem:V}
		Let $\{\hat{e}^b_k, \hat{e}^c_k\}_{k=1}^n$ be orthogonal vectors in $\mathbb{R}^m$
		whose components are defined by
		\begin{align*}
		[\hat{e}^b_k]_j &= \begin{cases}
		1 & \text{if } b_{k} < x_j \le b_{k+1}  \\
		0 & \text{elsewhere} 
		\end{cases},
		\qquad 
		[\hat{e}^c_k]_j 
		= \begin{cases}
		x_j - \mu_{k}  & \text{if }b_{k} < x_j \le b_{k+1} \\
		0 & \text{elsewhere} 
		\end{cases}.
		\end{align*}
		Then,  
		$
		V = \text{span}\{\xi^b_k, \xi^c_k\}_{k=1}^n = 
		\text{span}\{\hat{e}^b_k, \hat{e}^c_k\}_{k=1}^n
		$,
		and 
		$\text{rank}[V]  = 2n-2\bar{n}^{(0)} - \bar{n}^{(1)}$,
		where $\bar{n}^{(s)}$ is the number of entries of $\bm{u}=[u_1,\cdots, u_n]^T$ whose value is $s$.
	\end{lemma}
	\begin{proof}
		Let $V_{\xi} = \text{span}\{\xi^b_k, \xi^c_k\}_{k=1}^n$
		and $V_{e} = \text{span}\{\hat{e}^b_k, \hat{e}^c_k\}_{k=1}^n$.
		We first show that $V_{\xi} \subset V_{e}$.
		For each $k$, 
		$\xi^b_k = \sum_{j=k}^n \hat{e}^b_j$,
		and $\xi^c_k = \sum_{j=k}^n (\hat{e}^c_j + (\mu_{j} - b_k)\hat{e}^b_j)$.
		Also, note that 
		\begin{align*}
		\hat{e}^b_k = \xi^b_k - \xi^b_{k+1}, \qquad
		\hat{e}^c_k = \xi^c_k - \xi^c_{k+1} + (b_k-b_{k+1})\xi^b_{k+1} + (b_k - \mu_{k})(\xi^b_k - \xi^b_{k+1}),
		\end{align*}
		with $\xi_{n+1}^b = 0$ and $\xi_{n+1}^c = 0$,
		which shows $V_{\xi} \supset V_{e}$.
		Hence, $V_{\xi} = V_e$.
		
		The total number of vectors that span $V$ is $2n$.
		If $u_k = 0$, we have $\hat{e}^b_k = \hat{e}^c_k = 0$.
		If $u_k = 1$, $\hat{e}^c_k = 0$.
		Since $\{\hat{e}^b_j,\hat{e}^c_j\}_{j=1}^n$ are orthogonal,
		we conclude $\text{rank}[V] = 2n - 2\bar{n}^{(0)} - \bar{n}^{(1)}$.
	\end{proof}

	Let $V$ be the span of $\{\hat{e}^b_j,\hat{e}^c_j\}_{j=1}^n$
	and $\Pi_V[\text{Loss}]$ be
	the orthogonal projection of $\text{Loss}(t)$ onto $V$.
	For notational convenience, if $v$ is the zero vector, we simply think of 
	$\frac{v}{\|v\|}$ as the zero vector as well. 
	Let 
	$
	\bm{\Phi}_b = \begin{bmatrix}
	\hat{e}_1^b & \cdots & \hat{e}_n^b
	\end{bmatrix}$ and 
	$
	\bm{\Phi}_c = \begin{bmatrix}
	\hat{e}_1^c & \cdots & \hat{e}_n^c
	\end{bmatrix}
	$.
	It follows from
	$\xi^b_k = \sum_{j=k}^n \hat{e}^b_j$,
	and 
	$\xi^c_k = \sum_{j=k}^n (\hat{e}^c_j + (\mu_{j} - b_k)\hat{e}^b_j)$ 
	that $\bm{B}^T$ and $\bm{C}^T$ can be written as 
	\begin{align*}
	\bm{B}^T &= \begin{bmatrix}
	\xi_1^b & \cdots & \xi_n^b
	\end{bmatrix}D_c
	= \begin{bmatrix}
	\hat{e}_1^b & \cdots & \hat{e}_n^b
	\end{bmatrix}LD_c, \\
	\bm{C}^T &= \begin{bmatrix}
	\hat{e}_1^c & \cdots & \hat{e}_n^c
	\end{bmatrix}L
	+ \begin{bmatrix}
	\hat{e}_1^b & \cdots & \hat{e}_n^b
	\end{bmatrix}L',
	\end{align*}
	where $D_c$ is a diagonal matrix whose diagonal entries are $c_j$,
	$L$ and $L'$ are lower triangular matrices whose entries are defined by
	$$
	[L]_{ij} = \begin{cases}
	1 & \text{if } i \ge j \\
	0 & \text{elsewhere}
	\end{cases},\qquad
	[L']_{ij} = \begin{cases}
	\mu_{i} - b_j & \text{if } i \ge j \\
	0 & \text{elsewhere} 
	\end{cases}.
	$$
	Note that $L'(t)$ can be written as $D_{\mu}L - LD_b(t)$
	where $D_{\mu}$ and $D_b(t)$ are diagonal matrices whose $(i,i)$-components are defined to be $[D_{\mu}]_{ii} = \mu_i$
	and 
	$[D_b]_{ii} = b_i$, respectively.
	Also, let 
	$\bm{\Phi}_b = \bm{\bar{\Phi}}_b D_u$ and 
	$\bm{\Phi}_c = \bm{\bar{\Phi}}_c D_{\sigma}$,
	where
	$D_u$ and $D_{\sigma}$ are diagonal matrices whose $(i,i)$-components are defined respectively
	by
	$
	[D_u]_{ii} = \sqrt{u_i}$,
	and $[D_\sigma]_{ii} = \sqrt{u_i}\sigma_i$ for $i=1,\cdots, n$.
	By combining the above relations with \eqref{app:dt-loss}, we have
	\begin{align*}
	\frac{d}{dt}\mathcal{L}(t) &= \frac{1}{2}\frac{d}{dt}\|\text{Loss}(t)\|^2 
	= - \|\text{Loss}^T(t)
	\begin{bmatrix}
	\bm{B}^T & \bm{C}^T
	\end{bmatrix}\|^2
	\\
	&= - \|\text{Loss}^T(t)
	\begin{bmatrix}
	\bm{\Phi}_b & \bm{\Phi}_c
	\end{bmatrix}
	\begin{bmatrix}
	LD_c & L' \\
	\bm{0} & L
	\end{bmatrix}
	\|^2
	= - \|\text{Loss}^T(t)
	\begin{bmatrix}
	\bm{\bar{\Phi}}_b & \bm{\bar{\Phi}}_c
	\end{bmatrix}
	M(t)
	\|^2,
	\end{align*}
	where
	$$
	M(t) = 
	\begin{bmatrix}
	D_u & \bm{0} \\
	\bm{0} & D_{\sigma}
	\end{bmatrix}\begin{bmatrix}
	LD_c(t) & D_{\mu}L - LD_b(t) \\
	\bm{0} & L
	\end{bmatrix}.
	$$
	Note that during the time interval $[t_s, t_{s+1})$ on which $\{u_l(t)\}_{l=0}^n$ is constant,
	$D_u$, $D_\mu$, $D_{\sigma}$, $\bm{\bar{\Phi}}_b$, $\bm{\bar{\Phi}}_c$ and $L$
	are also constant.
	Thus, $b(t)$ and $c(t)$ determine the behavior of $M(t)$ where $t \in [t_s,t_{s+1})$. 
	Also, since the columns of $\bm{\bar{\Phi}}_b$ and $\bm{\bar{\Phi}}_c$ are orthonormal basis 
	that span $V$ defined in Lemma~\ref{app:lem:V},
	we have
	$
	\|\text{Loss}^T(t)
	\begin{bmatrix}
	{\bm{\bar{\Phi}}}_b & {\bm{\bar{\Phi}}}_c
	\end{bmatrix}
	\|^2
	=
	\|\Pi_V[\text{Loss}]\|^2$.
	For the interval $[t_s,t_{s+1})$ on which $\{u_k(t)\}$ is constant, 
	since $V_t = V_{t_s}$ for all $t \in [t_{s},t_{s+1})$,
	it can be checked that 
	$
	\frac{d}{dt}\mathcal{L}(t) = \frac{1}{2}\frac{d}{dt} \|\Pi_{V_t}[\text{Loss}(t)]\|^2$. 
	Let $\mathbf{r}_{\min}(t)$ and $\mathbf{r}_{\max}(t)$
	be the square of the ($\text{rank}(V_t)$)-th largest and the largest singular values of $M(t)$, respectively.
	Then, for $t \in [t_s, t_{s+1})$, we have
	\begin{equation*}
	-\mathbf{r}_{\max}(t)\|\Pi_{V_{t_s}}[\text{Loss}]\|^2
	\le 
	\frac{1}{2}\frac{d}{dt}\|\Pi_{V_{t_s}}[\text{Loss}]\|^2
	\le - \mathbf{r}_{\min}(t)\|\Pi_{V_{t_s}}[\text{Loss}]\|^2,
	\end{equation*}
	which implies 
	\begin{align*}
	\|\Pi_{V_t}[\text{Loss}(t)]\|^2
	&\le \|\Pi_{V_{t_s}}[\text{Loss}(t_s)]\|^2
	e^{-2\int_{t_s}^t \mathbf{r}_{\min}(v)dv}, \\
	\|\Pi_{V_t}[\text{Loss}(t)]\|^2
	&\ge 
	\|\Pi_{V_{t_s}}[\text{Loss}(t_s)]\|^2
	e^{-2\int_{t_s}^t \mathbf{r}_{\max}(v)dv},
	\end{align*}
	which completes the proof.
\end{proof}


\begin{lemma} \label{lem:def-M}
	Let 
	\begin{equation*}
	M(t) = 
	\begin{bmatrix}
	D_u & \bm{0} \\
	\bm{0} & D_{\sigma}
	\end{bmatrix}\begin{bmatrix}
	LD_c(t) & D_{\mu}L - LD_b(t) \\
	\bm{0} & L
	\end{bmatrix} \in \mathbb{R}^{2n\times 2n},
	\end{equation*}
	where $L$ is an upper triangular matrix that is independent of $t$ 
	whose $(i,j)$-component is $1$ for $i \le j$, and 0 otherwise,
	and 
	$D_u, D_\mu, D_\sigma, D_c, D_b$ 
	are diagonal matrices of size $n$ 
	whose $(i,i)$-components are 
	defined to be 
	$$
	[D_u]_{ii} = \sqrt{u_i}, \quad
	[D_\mu]_{ii} = \mu_i, \quad 
	[D_\sigma]_{ii} = \sqrt{u_i}\sigma_i, \quad 
	[D_c]_{ii} = c_i, \quad 
	[D_b]_{ii} = b_i,
	$$
	respectively, for $i=1,\cdots,n$,
	where $\mu_i, \sigma_i$ are defined in \eqref{def:mean-variance-data}.
	Let $V_t$ be the space defined in Lemma~\ref{app:lem:V}.
	Then, 
	the square of the $\text{rank}(V_t)$-th largest $\mathbf{r}_{\min}(t)$ and the largest $\mathbf{r}_{\max}(t)$
	singular values of the matrix $M(t)$ satisfy
	\begin{align*}
	\mathbf{r}_{\min}(t) &\ge \frac{\min_l\{u_l\min\{1,\sigma^2_{l}\}\}}{4(1 + (\min_l|c_{l}^2(t)|)^{-1} + 4d_{\max}^2n^2)},
	\\
	\mathbf{r}_{\max}(t) &\le \max_l\{u_l\max\{1,\sigma^2_{l}\}\}(1 + \max_l|c_{l}^2(t)| + d_{\max}^2)n^2,
	\end{align*}
	where
	$d_{\max} = x_m - x_1$
	and
	the minima are taken over $\{l \in [n] | u_l \ne 0, \sigma_{l} \ne 0\}$.
\end{lemma}
\begin{proof}
	Recall that
	\begin{equation} \label{def:M-mat-2}
	M = 
	DN,
	\qquad
	D = \begin{bmatrix}
	D_u & \bm{0} \\
	\bm{0} & D_{\sigma}
	\end{bmatrix},
	\quad
	N = \begin{bmatrix}
	LD_c(t) & D_{\mu}L - LD_b(t) \\
	\bm{0} & L
	\end{bmatrix}.
	\end{equation}
	From now on, all the minima are taken over all $k$ but $u_k=0$ and $\sigma_{k} = 0$.
	Since $D$ is diagonal,
	the smallest nonzero $\sigma_{\min}(D)$ and the largest $\sigma_{\max}(D)$ singular values of $D$ 
	satisfy
	$
	\sigma_{\min}^2(D) = \min_k\{u_k\min\{1,\sigma^2_{k}\}\}$ and 
	$\sigma_{\max}^2(D) = \max_k\{u_k\max\{1,\sigma^2_{k}\}\}$.

	The inverse $N^{-1}$ of $N$ is given by 
	$$
	N^{-1} = \begin{bmatrix}
	LD_c & L' \\
	\bm{0} & L
	\end{bmatrix}^{-1} = \begin{bmatrix}
	D_c^{-1}L^{-1} & -D_c^{-1}L^{-1}L'L^{-1} \\ \bm{0} & L^{-1}
	\end{bmatrix},
	$$
	and the following holds:
	\begin{align*}
	\|N\|^2 &\le \|LD_c\|^2 + \|L\|^2 + \|L'\|^2,
	\\ 
	\|N^{-1}\|^2 &\le \|D_c^{-1}L^{-1}\|^2 + \|L^{-1}\|^2 + \|D_c^{-1}L^{-1}L'L^{-1}\|^2.
	\end{align*}
	It can also be checked that $L^{-1}$ is an lower triangular matrix such that 
	$[L^{-1}]_{ij} = 1$ if $i=j$, 
	$[L^{-1}]_{ij} = -1$ if $i=j+1$
	and zero elsewhere.
%
	Recall that for a matrix $A$, we have 
	$
	\|A\|_2 \le \sqrt{\|A\|_1\|A\|_\infty}$.
	Since $\|L\|_1 = \|L\|_\infty = n$, we have $\|L\|\le n$.
	Similarly, we have $\|L^{-1}\| \le 2$.
	Also, note that 
	\begin{align*}
	\|LD_c\|^2 &\le 
	\left(\max_j (n-j+1)|c_j(t)|\right)
	\sum_{k=1}^n |c_k(t)|, \\
	\|D_c^{-1}L^{-1}\|^2 &\le
	\max\{|c_1(t)|^{-1}, 2\max_{1<j\le n} |c_j(t)|^{-1}\}\cdot 
	\max_{1\le j<n} (|c_j(t)|^{-1} + |c_{j+1}(t)|^{-1}).
	\end{align*}
	Thus, it suffices to compute $\|L'\|$.
	Note that 
	$
	\|L'\|_1 = \sum_{j=1}^n (\mu_{U_j}-b_1)$,
	and $\|L'\|_\infty = \sum_{j=1}^n (\mu_{U_n}-b_j)$.
	Let $d_{\max} = \max_j x_j - \min_j x_j$. 
	Then, $\|L'\| \le d_{\max}n$.
	Thus,
	\begin{align*}
	\|N\|^2 \le (|c_{\max}(t)|^2 + 1 + d_{\max}^2)n^2,
	\qquad
	\|N^{-1}\|^2 \le 4(1 + |c_{\min}(t)|^{-2} + 4d_{\max}^2n^2),
	\end{align*}
	where $c_{\min}(t) = \min_k |c_k(t)|$
	and $c_{\max}(t) = \max_k |c_k(t)|$.
	
	Hence, we obtain
	\begin{align*}
	\sigma_{\min}^2(M(t)) &\le \frac{\min_k\{u_k\min\{1,\sigma^2_{k}\}\}}{4(1 + |c_{\min}^2(t)|^{-1} + 4d_{\max}^2n^2)}, \\
	\sigma_{\max}^2(M(t)) &\ge -\max_k\{u_k\max\{1,\sigma^2_{k}\}\}(|c_{\max}^2(t)| + 1 + d_{\max}^2)n^2,
	\end{align*}
	which completes the proof.
\end{proof}

\section{Proof of Lemma~\ref{lemma:critical}} \label{app:lemma:critical}
\begin{proof}
%
	Let 
	$
	\Omega^* = \{(\bm{b}, \bm{c}) \in \mathbb{R}^n \times \mathbb{R}^n  | \bm{M}_b(\bm{c})\bm{b}  = \bm{g}(\bm{c}), \bm{M}_c(\bm{b})\bm{c}  = \bm{C}(\bm{b})\bm{y} \}$.
	Without loss of generality, let us assume that each component of $\bm{c}^*$ is nonzero.
	If $\bm{c}^*$ has a zero entry, its corresponding neuron can be ignored in all the arguments.
	Note that 
	\begin{align*}
	&\bm{M}_b(\bm{c}^*)\bm{b}^*  = \bm{g}(\bm{c}^*)
	\iff D_{c^*}L^TD_u^2 L D_{c^*} \bm{b}= D_{c^*}L^TD_u^2(D_\mu L\bm{c}^*- D_u^{-1}\bar{\Phi}_b^T\bm{y}).
	\\
	&\iff 
	D_u^2 L D_{c^*} \bm{b}= D_u^2(D_\mu L\bm{c}^*- D_u^{-1}\bar{\Phi}_b^T\bm{y}) \\
	&\iff 
	D_u^2(D_\mu L - LD_b)\bm{c}^* = D_u\bar{\Phi}_b^T\bm{y}
	\iff
	D_u^2 L_{\bm{b}^*}'\bm{c}^* = 	\Phi_b^T\bm{y}.
	\end{align*}
	This implies that 
	\begin{align*}
	D_u^2 L D_{c^*} \bm{b}= D_u^2(D_\mu L\bm{c}^*- D_u^{-1}\bar{\Phi}_b^T\bm{y})
	\iff
	\sum_{i=1}^{k} c^*_i (\mu_{k} - b^*_i) = \bar{y}_{k}, 
	\end{align*}
	for all $k$ such that $u_k > 0$,
	where $\bar{y}_k$ and $\mu_k$ are defined in \eqref{def:mean-variance-data}.
	Therefore, this shows that 
	any critical points lead a piecewise linear interpolation of the macroscopic data
	$\{(\mu_{k},\bar{y}_{k})\}_{k \in \{j|u_j > 0\}}$.
	Also, we note that 
	\begin{align*}
	&\bm{M}_c(\bm{b}^*)\bm{c}^*  = \bm{C}(\bm{b}^*)\bm{y}
	\iff 
	((L_{\bm{b}^*}')^TD_u^2L_{\bm{b}^*}' + L^T D_x^2 L) \bm{c}^* = (L'_{\bm{b}^*})^T\Phi_b^T\bm{y} + L^T\Phi_c^T\bm{y}
	\\
	&\iff 
	L^T D_x^2 L \bm{c}^* = L^T\Phi_c^T\bm{y}
	\iff 
	D_x^2 L \bm{c}^* = \Phi_c^T\bm{y},
	\end{align*}
	which can be equivalently written as follows:
	For any $k$ such that $u_k > 1$,
	$
	\sum_{i=1}^{k} c_i^* = 
	\frac{\frac{1}{u_k}\sum_{(x,y) \in U_k} xy - \mu_k\bar{y}_k}{\sigma^2_{k}}$.
	Therefore, any pair $(\bm{b}^*,\bm{c}^*)$ satisfying 
	$D_u^2 L_{\bm{b}^*}'\bm{c}^* = 	\Phi_b^T\bm{y}$
	and
	$D_x^2 L \bm{c}^* = \Phi_c^T\bm{y}$
	is a critical point of the gradient flow dynamics \eqref{def:bias-gradflow}.
	Indeed, this is solvable. 
	For a square matrix $A$, let $A^\dagger$ be its Moore-Penrose inverse.
	Then, 
	$\bm{c}^* = (D_x^2L)^\dagger \Phi_c^T\bm{y}$.
	Since $L_{\bm{b}^*} = D_{\mu}L - LD_{\bm{b}^*}$, we have
	\begin{align*}
	D_u^2 L_{\bm{b}^*}'\bm{c}^* = 	\Phi_b^T\bm{y} 
	&\iff
    D_u^2 (D_{\mu}L - LD_{\bm{b}^*})\bm{c}^* = 	\Phi_b^T\bm{y}  \\
    &\iff
    D_u^2LD_{\bm{c}^*}\bm{b}^* = D_u^2D_{\mu}L\bm{c}^* - \Phi_b^T\bm{y} \\
    &\impliedby
    \bm{b}^* = (D_u^2LD_{\bm{c}^*})^\dagger (D_u^2D_{\mu}L\bm{c}^* - \Phi_b^T\bm{y}).
	\end{align*}
	Thus, $\Omega^*$ is not empty and the proof is completed.
\end{proof}

\section{Proof of Theorem~\ref{thm:convergence}} \label{app:thm:convergence}
\begin{proof}
	Let $V_{t_s} = \text{span}\{\hat{e}^b_l(t_s), \hat{e}^c_l(t_s) \}_{l=1}^n$.
	Since $\{u_l\}$ remains constant for $[t_s,\infty)$
	and $c_j$ does not vanish at infinite for all $j$ with $u_j \ne 0$,
	it then follows from Theorem~\ref{thm:conti-loss-n} that for sufficiently large $t$,
	$
	\|\Pi_{V_{t_s}} [\text{Loss}(t)]\| \le Ce^{-\gamma t}$,
	where $C, \gamma > 0$ are some positive constants.
	Observe that
	\begin{align*}
	\|\Pi_{V_{t_s}} [\text{Loss}(t)]\| &= 
	\|\bar{\bm{\Phi}}^T
	[\bm{C}^T(t)\bm{c} - \bm{y}]\|
	= \|\begin{bmatrix}
	D_u(D_\mu L - LD_b) \\ D_\sigma L
	\end{bmatrix}\bm{c}
	- \bar{\bm{\Phi}}^T\bm{y}\|,
	\end{align*}
	where $\bar{\bm{\Phi}} = [\bar{\bm{\Phi}}_b, \bar{\bm{\Phi}}_c]$.
	Hence, in the limit, since $\|\Pi_{V_{t_s}} [\text{Loss}(t)]\| \to 0$, we have
	\begin{align*}
	\lim_{t\to \infty} D_u(D_\mu L - LD_b(t))\bm{c}(t) = \bm{\bar{\Phi}}_b^T\bm{y},
	\qquad
	\lim_{t\to \infty}  D_\sigma L\bm{c}(t) = \bm{\bar{\Phi}}_c^T\bm{y}.
	\end{align*}
	Also, note that
	$$
	\begin{bmatrix}
	\dot{\bm{b}}(t) \\
	-\dot{\bm{c}}(t)
	\end{bmatrix}
	=
	M^T(t) \bar{\bm{\Phi}}^T\text{Loss}(t) = 
	M^T(t) \left(\begin{bmatrix}
	D_u(D_\mu L - LD_b) \\ D_\sigma L
	\end{bmatrix}\bm{c}
	- \bar{\bm{\Phi}}^T\bm{y} \right),
	$$
	where $M(t)$ is defined in \eqref{def:M-mat-2}.
	It follows from 
	\begin{align*}
	\|\text{Loss}(0)\|^2 &\ge   
	-\int_{0}^s \frac{d}{dt} \|\text{Loss}(t)\|^2 dt 
	= 2\int_0^s \left(\|\dot{\bm{b}}(t)\|^2 + \|\dot{\bm{c}}(t)\|^2 \right) dt
	\\
	&\ge 2\left(\|\bm{b}(t) - \bm{b}(0)\|^2 + \|\bm{c}(t) - \bm{c}(0)\|^2 \right).
	\end{align*}
	that $\|\bm{c}(0) - \bm{c}(t)\|^2 + \|\bm{b}(0) - \bm{b}(t)\|^2 \le \mathcal{L}(0)$ for all $t \ge 0$, which shows the uniformly boundedness of $\bm{b}(t), \bm{c}(t)$.
	Therefore, by Lemma~\ref{lem:def-M}, $\|M(t)\|$ is also uniformly bounded
	so that one can conclude that $\lim_{t\to \infty} \dot{\bm{b}}(t) = \lim_{t\to \infty} \dot{\bm{c}}(t) = \bm{0}$.
	
	We now show that 
	$\lim_{t\to \infty} \bm{b}(t) = \bm{b}^*$
	and $\lim_{t\to \infty} \bm{c}(t) = \bm{c}^*$
	for some $\bm{b}^*, \bm{c}^* \in \mathbb{R}^n$.
	We proceed the proof by focusing on $\bm{b}(t)$. The proof of $\bm{c}(t)$ follows similarly.
	Note that for sufficiently large $t$, we have 
	$$
	\|\dot{\bm{b}}(t)\| \le \left(\sup_t \|M(t)\|\right)\|\Pi_{V_{t_s}} [\text{Loss}(t)]\| \le Ce^{-\gamma t}.
	$$
	For sufficiently large $p$, by the mean value theorem, there exists $t_p \in [p,p+1]$ such that 
	$
	\|\bm{b}(p+1) - \bm{b}(p)\|\le \|\dot{\bm{b}}(t_p)\| \le Ce^{-\gamma p}$. 
	Let us consider the sequence $\{\bm{b}(p)\}_{p=1}^{\infty}$.
	We first show that the sequence is Cauchy.
	Note that for any $m, n \in \mathbb{N}$ (let $m > n$), we have
	\begin{align*}
	\|\bm{b}(m) - \bm{b}(n)\| &= \| \sum_{j=n}^{m-1} \bm{b}(j+1) - \bm{b}(j)\|
	\le \sum_{j=n}^{m-1} \|\bm{b}(j+1) - \bm{b}(j)\|
	\\
	&\le C \sum_{j=n}^{m-1} e^{-\gamma j} = C\cdot \frac{e^{-\gamma n} - e^{-\gamma m}}{1 - e^{-\gamma}}.
	\end{align*}
	Therefore, $\{\bm{b}(p)\}_{p=1}^{\infty}$ is Cauchy and let $\bm{b}^*$ be its limit point.

	We now prove $\lim_{t\to \infty} \bm{b}(t) = \bm{b}^*$ by contradiction.
	Suppose $\lim_{t\to \infty} \bm{b}(t) \ne \bm{b}^*$.
	Then, there exists $\epsilon > 0$ such that for any $T$
	there exists $s > T$ such that $\|\bm{b}(s) - \bm{b}^*\| > \epsilon$.
	Since $\lim_{p\to \infty} \bm{b}(p) = \bm{b}^*$,
	there exists $K$ such that for all $p \ge K$, $\|\bm{b}(p) - \bm{b}^*\| < \epsilon/2$.
	Then, for sufficiently large $k > K$, there exists $s > k$ such that 
	\begin{align*}
	\|\bm{b}(s) - \bm{b}(k)\| \ge 
	\|\bm{b}(s) - \bm{b}^*\| 
	- \|\bm{b}(k)- \bm{b}^*\| > \epsilon/2.
	\end{align*}
	Let $k'$ be the closest integer to $s$.
	Then since $|k' -s | \le 1$ and $k \le s, k'$, we have 
	\begin{align*}
	\|\bm{b}(k) - \bm{b}(s)\| &\le 
	\|\bm{b}(k) - \bm{b}(k')\| + \|\bm{b}(k') - \bm{b}(s)\| 
	\le C\cdot \frac{e^{-\gamma k}}{1 - e^{-\gamma}} + Ce^{-\gamma k}.
	\end{align*}
	Let $k$ be sufficiently large to satisfy
	$
	C\cdot \frac{e^{-\gamma k}}{1 - e^{-\gamma}} + Ce^{-\gamma k} < \frac{\epsilon}{4}
	$.
	Then, we have
	$
	\frac{\epsilon}{2} < \|\bm{b}(k) - \bm{b}(s)\|  < \frac{\epsilon}{4},
	$
	which is a contradiction.
	Hence, we conclude that $\lim_{t\to \infty} \bm{b}(t) = \bm{b}^*$.
\end{proof}

\section{Proof of Theorem~\ref{thm:LSQ-Stationary}} \label{app:thm:LSQ-Stationary}

\begin{lemma} \label{app:lem:LSQ}
	Suppose there are $m$-data points $\{(x_j, y_j)\}_{j=1}^{m}$.
	Let $\bar{x} = \frac{1}{m}\sum_{j=1}^{m} x_j$, $\bar{y} = \frac{1}{m}\sum_{j=1}^{m} y_j$, $\overline{x^2} = \frac{1}{m}\sum_{j=1}^{m} x_j^2$
	and $\text{Var}[X] = \overline{x^2} - (\bar{x})^2$.
	
	For $m\ge 2$, the least square solution line is given by
	\begin{equation*}
	l_{\text{best}}(x) = \left(\frac{\frac{1}{m}\sum_{j=1}^{m}x_jy_j  - \bar{x} \bar{y}}{\text{Var}[X]}\right)x + \frac{\overline{x^2}\bar{y} -\frac{\bar{x}}{m}\sum_{i=1}^m x_i y_i}{\text{Var}[X]}.
	\end{equation*}
	Also, $l_{\text{best}}(\bar{x}) = \bar{y}$ is satisfied.	
	
	For $m=1$, the line that interpolates a single datum is the form of 
	\begin{equation*}
	l(x;\alpha) = \frac{x_1y_1}{x_1^2+1}x + \frac{y_1}{x_1^2+1} + \alpha(x-x_1),
	\qquad \alpha \in \mathbb{R}.
	\end{equation*}
	Also, $l(x_1;\alpha) = y_1$ is satisfied.	
	When $\alpha=0$, it is the least norm solution.
\end{lemma}
\begin{proof}
	It can be checked that the least square solution $(c^*, d^*)$ satisfies 
	\begin{align*}
	\begin{bmatrix}
	\sum_{i=1}^m x_i^2 &  \sum_{i=1}^m x_i \\
	\sum_{i=1}^m x_i & m
	\end{bmatrix}
	\begin{bmatrix}
	c \\ d
	\end{bmatrix}
	=
	\begin{bmatrix}
	\sum_{i=1}^m x_i y_i \\ \sum_{i=1}^n y_i
	\end{bmatrix}.
	\end{align*}
	When $m \ge 2$, 
	the solution is given by
	\begin{equation*}
	\begin{bmatrix}
	c^* \\ d^*
	\end{bmatrix}
	= \frac{1}{\text{Var}[X]}\begin{bmatrix}
	1 & -\bar{x} \\ -\bar{x} & \overline{x^2} 
	\end{bmatrix}
	\begin{bmatrix}
	\frac{1}{m}\sum_{i=1}^m x_i y_i \\ \bar{y}
	\end{bmatrix}
	=
	\frac{1}{\text{Var}[X]}\begin{bmatrix}
	\frac{1}{m}\sum_{i=1}^m x_i y_i - \bar{x}\bar{y}
	\\
	\overline{x^2}\bar{y} -\frac{\bar{x}}{m}\sum_{i=1}^m x_i y_i
	\end{bmatrix}.
	\end{equation*}
	The relation of $c^*\bar{x} + d^* = \bar{y}$ can be readily checked.
	When $m=1$, any solution that interpolates a single point is given by
	$
	l(x;\alpha) = \frac{x_1y_1}{x_1^2+1}x + \frac{y_1}{x_1^2+1} + \alpha(x-x_1)
	$,
	for any $\alpha \in \mathbb{R}$.
	When $\alpha=0$, $l(x;0)$ is the least norm solution. 
\end{proof}
%

\begin{lemma} \label{app:lem:uk=2}
	Suppose that there exists $t_0 \ge 0$
	such that for any $t \in [t_0, \infty)$, 
	$\{u_k(t)\}$ is constant.
	If $u_k(t) \ge 2$ for all $k$, 
	the gradient flow dynamics \eqref{def:bias-gradflow} has a unique 
	stationary point $(\bm{b}^*, \bm{c}^*)$.
	Furthermore, for each $k$, 
	$(b^*_k, c^*_k)$ is the least square solution 
	on the training data $\{(x_s, \hat{y}_s)\}_{s=1}^{u_k}$ 
	where $x_s \in (b^*_k, b^*_{k+1}]$ and
	$\hat{y}_s = y_s - \sum_{j=1}^{k-1} c_j^*\phi(x_s - b_j^*)$.
\end{lemma}
\begin{proof}
	Let $U_i$ be the index set of training input data in the interval 
	$(b_i, b_{i+1}]$.
	Let $\bm{x}_i$ be the vector of size $u_i$ whose entries are the training input data corresponding to $U_i$.
	Similarly, $\bm{y}_i$ is defined.
	Let $\overline{\langle \bm{x}_i, \bm{y}_i\rangle } := \frac{1}{u_i}\langle \bm{x}_i, \bm{y}_i\rangle$ and $\bar{\bm{y}}_i$ be the mean of $\bm{y}_i$.
	Also, let $\bar{\bm{x}}_i$ and $\text{Var}[\bm{x}_i]$
	be the mean and the variance of $\bm{x}_i$, respectively.
	
	Since $u_k \ge 2$ for all $k$, 
	the uniqueness of the stationary point follows from
	Lemma~\ref{lemma:critical}.
	Let $(\bm{b}^*,\bm{c}^*)$ be the stationary point
	and let us write the resulting ReLU approximation by
	$
	f^*(x) = \sum_{i=1}^n c^*_i \phi(x - b^*_i)$, where $b_1^* \le \cdots \le b_n^*$.
	Since $u_1 \ge 2$, in the first interval $(b_1,b_2]$, 
	it follows from Lemma~\ref{lemma:critical} that 
	we have
	$
	c_1^* = \frac{\overline{\langle \bm{x}_i, \bm{y}_i\rangle} - \bar{\bm{x}}_i\bar{\bm{y}}_i}{\text{Var}[\bm{x}_i]}$,
	$b_1^* = \bar{\bm{x}}_1 - \frac{\bar{\bm{y}}_1}{c_1^*}$.
	From Lemma~\ref{app:lem:LSQ}, it can be checked that $(b_1^*, c_1^*)$ is identical to the least square solution using the data $(\bm{x}_1,\bm{y}_1)$.
	
	For a vector $\bm{v}$, the $i$-th component of $\bm{v}$ is denoted by $[\bm{v}]_i$.
	Suppose the statement is true for all $k < s$, i.e., 
	$(b_k^*, c_k^*)$ is the least square solution using the training data
	$(\bm{x}_k, \hat{\bm{y}}_i)$, where 
	$
	[\hat{\bm{y}}_i]_t = [\bm{y}_i]_t - \sum_{j=1}^{k-1} c_j^*\phi([\bm{x}_k]_t - b_j^*),  \forall 1 \le t \le u_k.
	$
	Let $f_k(\bm{x}_k)$ be a vector of size $n_k$ whose $t$-th entry is
	$\sum_{j=1}^{k-1} c_j^*\phi([\bm{x}_k]_t - b_j^*)$.
	Then, one can concisely write $\hat{\bm{y}}_k = \bm{y}_k - f_k(\bm{x}_k)$.
	
	When $k=s$, it follows from Lemma~\ref{lemma:critical} that 
	we have
	$
	c_s^* = \frac{\overline{\langle \bm{x}_s, \bm{y}_s\rangle} - \bar{\bm{x}}_s\bar{\bm{y}}_s}{\text{Var}[\bm{x}_s]} - \frac{\overline{\langle \bm{x}_{s-1}, \bm{y}_{s-1}\rangle} - \bar{\bm{x}}_{s-1}\bar{\bm{y}}_{s-1}}{\text{Var}[\bm{x}_{s-1}]}
	$,
	$b_s^* = \bar{\bm{x}}_s - \frac{\bar{\bm{y}}_s}{c_s^*}$.
	Note that 
	\begin{align*}
	\frac{\overline{\langle \bm{x}_s, \hat{\bm{y}}_s\rangle} - \bar{\bm{x}}_s\bar{\hat{\bm{y}}}_s}{\text{Var}[\bm{x}_s]}
	&=\frac{\overline{\langle \bm{x}_s, \bm{y}_s\rangle} - \bar{\bm{x}}_s\bar{\bm{y}}_s}{\text{Var}[\bm{x}_s]}
	-\frac{\overline{\langle \bm{x}_s, f_s(\bm{x}_s)\rangle} - \bar{\bm{x}}_s\overline{f_s(\bm{x}_s)}}{\text{Var}[\bm{x}_s]}.
	\end{align*}
	Since 
	\begin{align*}
	\overline{\langle \bm{x}_s, f_s(\bm{x}_s)\rangle}
	= \sum_{j=1}^{s-1} c_j^* \overline{\langle \bm{x}_s, \bm{x}_s\rangle} 
	-\sum_{j=1}^{s-1} c_j^*b_j^* \bar{\bm{x}}_s, \quad
	\bar{\bm{x}}_s\overline{f_s(\bm{x}_s)}
	= (\bar{\bm{x}}_s)^2 \sum_{j=1}^{s-1}c_j^* - \bar{\bm{x}}_s\sum_{j=1}^{s-1}c_j^*b_j^*,
	\end{align*}
	we have
	\begin{align*}
	\frac{\overline{\langle \bm{x}_s, f_s(\bm{x}_s)\rangle} - \bar{\bm{x}}_s\overline{f_s(\bm{x}_s)}}{\text{Var}[\bm{x}_s]}
	&= \frac{\left[\overline{\langle \bm{x}_s, \bm{x}_s\rangle}  -(\bar{\bm{x}}_s)^2\right] \sum_{j=1}^{s-1}c_j^*}{\text{Var}[\bm{x}_s]}
	= \sum_{j=1}^{s-1}c_j^*.
	\end{align*}
	Therefore, $(b_s^*, c_s^*)$ is the least square solution using the training data
	$(\bm{x}_s,\hat{\bm{y}}_s)$.
	By induction, the proof is completed.
\end{proof}

\begin{proof}[Proof of Theorem~\ref{thm:LSQ-Stationary}]
	From Lemma~\ref{app:lem:LSQ} and ~\ref{app:lem:uk=2},
	it suffices to consider the case where $u_k = 1$.
	It follows from Lemma~\ref{lemma:critical}
	that the ReLU network has to interpolate $(x_1^k, y_1^k)$.
	Hence, the proof is completed by applying Lemma~\ref{app:lem:LSQ}.
\end{proof}

\section{Proof of Theorem~\ref{thm:Q-decrease}} \label{app:thm:Q-decrease}

Let $[t_s,t_{s+1})$ be the time interval on which 
$\{u_l(t)\}_{l=1}^n$ is constant.
Suppose that at time $t_{s+1}$, there exists $j$ such that $u_j(t_s) = u_j(t_{s+1}) \pm 1$,
$u_{j+1}(t_s) = u_{j+1}(t_{s+1}) \mp 1$,
and $u_l(t_s) = u_l(t_{s+1})$ for all $l$ but $j$ and $j+1$.
We note that this happens only when $b_{j+1}$ crosses the largest data point in $U_j(t_s)$
(the smallest data point in $U_{j+1}(t_s)$)
from the right (the left) at time $t_{s+1}$.  

Let $V_t^{\text{inv}} = \text{span}\{\hat{e}^b_l(t), \hat{e}^c_l(t) : l \ne j, j+1\}$
and $V_t^{\text{var}} =  \text{span}\{\hat{e}^b_l(t), \hat{e}^c_l(t)\}_{l=j}^{j+1}$.
Then $V_{t} = V_t^{\text{inv}} \oplus V_t^{\text{var}}$.
Since $V_{t_s}^{\text{inv}} =  V_{t_{s+1}}^{\text{inv}}$,
we denote it as $V^{\text{inv}}$.
Also, note that $V_{t_s}^{\text{var}} \ne V_{t_{s+1}}^{\text{var}}$.

\begin{lemma} \label{app:lem:e-vectors}
	There exists two vectors $\hat{e}^n(t_s) \in V_{t_s}^\perp$
	and $\hat{e}^n(t_{s+1}) \in V_{t_{s+1}}^\perp$ such that 
	$$
	V^{\text{mix}}:=V_{t_s}^{\text{var}} \oplus \hat{e}^n(t_{s}) = V_{t_{s+1}}^{\text{var}} \oplus \hat{e}^n(t_{s+1}),
	\qquad
	V^{\text{null}}:=V_{t_s}^\perp\backslash \hat{e}^n(t_s)= V_{t_{s+1}}^\perp\backslash \hat{e}^n(t_{s+1}).
	$$
	For the later use, let us define $\zeta = \frac{\hat{e}^n(t_{s})}{\|\hat{e}^n(t_{s})\|}$ and $\psi = \frac{\hat{e}^n(t_{s+1})}{\|\hat{e}^n(t_{s+1})\|}$.
\end{lemma}
\begin{proof}
	Without loss of generality, let us assume that $u_j(t_s) = u_j(t_{s+1}) + 1$,
	$u_{j+1}(t_s) = u_{j+1}(t_{s+1}) - 1$.

	We observe that $\hat{e}^b_{j+1}(t_{s+1}) - \hat{e}^b_{j+1}(t_s) = \hat{e}^b_j(t_s) - \hat{e}^b_{j}(t_{s+1}):=e^n$.
	Let $V_t^\text{mix} = \text{span}\{\hat{e}^b_{j}(t), \hat{e}^b_{j+1}(t), \hat{e}^c_{j}(t), \hat{e}^c_{j+1}(t), e^n\}$.
	It then follows from 
	\begin{align*}
	\hat{e}^c_j(t_s) - \hat{e}^c_{j}(t_{s+1}) &\in 
	\text{span}\{\hat{e}^b_{j}(t_{s+1}), e^n\}
	=\text{span}\{\hat{e}^b_{j}(t_{s}), e^n\}, 
	\\
	\hat{e}^c_{j+1}(t_s) - \hat{e}^c_{j+1}(t_{s+1}) &\in \text{span}\{\hat{e}^b_{j+1}(t_{s}), e^n\}
	=
	\text{span}\{\hat{e}^b_{j+1}(t_{s+1}), e^n\},
	\end{align*}
	that 
	$V_{t_s}^\text{mix} = V_{t_{s+1}}^{\text{mix}}$.
	Since $\{\hat{e}^b_{l}(t), \hat{e}^c_{l}(t)\}_{l=j}^{j+1}$ is orthogonal,
	by applying an orthogonalization procedure (e.g. Gram-Schimidtz), 
	one obtain an orthogonal basis of $V_t^{\text{mix}}$, $\{\hat{e}^b_{l}(t), \hat{e}^c_{l}(t)\}_{l=j}^{j+1} \cup \{\hat{e}^n(t)\}$, where 
	$$
	\hat{e}^n(t) = e^n(t) - \sum_{l=j}^{j+1} \left[\frac{\langle e^n(t), \hat{e}_l^b(t)\rangle }{\|\hat{e}_l^b(t)\|^2}\hat{e}_l^b(t)
	+ \frac{\langle e^n(t), \hat{e}_l^c(t)\rangle }{\|\hat{e}_l^c(t)\|^2}\hat{e}_l^c(t)\right].
	$$
	Also, since the support of $e^n(t)$ is disjoint to those of $V_t^\text{inv}$,
	we have $e^n(t) \in V_t^\perp$, which completes the proof.
\end{proof}

For any $u \in \mathbb{R}^m$, let 
$\Pi_V$ be the orthogonal projection of $u$ onto the space $V$.
Let $V_t^n = \text{span}\{\hat{e}^n(t)\}$.
Let
$$
u_1 = \frac{\hat{e}^b_j(t_s)}{\|\hat{e}^b_j(t_s)\|}, u_2=\frac{\hat{e}^b_{j+1}(t_s)}{\|\hat{e}^b_{j+1}(t_s)\|},
u_3=\frac{\hat{e}^c_j(t_s)}{\|\hat{e}^c_j(t_s)\|},
u_4=\frac{\hat{e}^c_{j+1}(t_s)}{\|\hat{e}^c_{j+1}(t_s)\|}, 
u_5=\frac{\hat{e}^n(t_s)}{\|\hat{e}^n(t_s)\|},
$$
and $v^* = \frac{\hat{e}^n(t_{s+1})}{\|\hat{e}^n(t_{s+1})\|}$.
It then can be checked that 
\begin{align*}
Q_0(t_s) &= \|\Pi_{V_{t_s}^\perp} \text{Loss}(t_s)\|^2 
=  \|\Pi_{V^{\text{null}}} \text{Loss}(t_s)\|^2 
+ \|\Pi_{V_{t_{s}}^n} \text{Loss}(t_s)\|^2, \\
Q_0(t_{s+1}) &= \|\Pi_{V_{t_{s+1}}^\perp} \text{Loss}(t_{s+1})\|^2 
=  \|\Pi_{V^{\text{null}}} \text{Loss}(t_{s+1})\|^2 
+ \|\Pi_{V_{t_{s+1}}^n} \text{Loss}(t_{s+1})\|^2.
\end{align*}
Since $\|\Pi_{V^{\text{null}}} \text{Loss}(t_s)\|^2 =\|\Pi_{V^{\text{null}}} \text{Loss}(t_{s+1})\|^2$, we have
$$
Q_0(t_s) - Q_0(t_{s+1}) = \|\Pi_{V_{t_{s}}^n} \text{Loss}(t_s)\|^2 - \|\Pi_{V_{t_{s+1}}^n} \text{Loss}(t_{s+1})\|^2.
$$
Since $\hat{e}^n(t_{s+1}) \in V^{\text{mix}}$
can be written as a linear combination of $\{u_i\}_{i=1}^5$,
we obtain
\begin{align*}
	\|\Pi_{V_{t_{s+1}}^n} \text{Loss}(t)\|^2
	= |\langle \text{Loss}(t), v^* \rangle|^2 
	= \left(\sum_{i=1}^5 u_i^Tv^*   \langle \text{Loss}(t), u_i \rangle \right)^2.
\end{align*}
Note that for $i=1,2,3,4$, 
$|\langle \text{Loss}(t), u_i \rangle|$ continuously decreases
as $t$ increases in $[t_s,t_{s+1})$.
Also, if $t_{s+1} = \infty$,  $|\langle \text{Loss}(t), u_i \rangle| \to 0$ as $t \to \infty$ for $i=1,2,3,4$.
Note also that $|u_5^Tv^*| < 1$.
Suppose 
there exists $t^* \in [t_s,t_{s+1}]$ such that 
for all $t \in [t^*,t_{s+1}]$,  
\begin{equation} \label{app:Q-decay-cond}
-1-s_5u_5^Tv^*
<
\frac{\left\langle \text{Loss}(t), \Pi_{V_{t_s}^{\text{var}}}[v^*]\right\rangle }{|\langle \text{Loss}(t_s), u_5 \rangle|}
< 1-s_5u_5^Tv^*,
\end{equation}
where $s_5 = \text{sign}(\langle \text{Loss}(t_s), u_5 \rangle)$.
Then, we can conclude that 
$$
\|\Pi_{\hat{e}^n(t_{s+1})} \text{Loss}(t_{s+1})\|^2 < |\langle \text{Loss}(t_s), u_5 \rangle|^2 = \|\Pi_{\hat{e}^n(t_s)} \text{Loss}(t_s)\|^2,
$$
which is equivalent to $Q_0(t_s) > Q_0(t_{s+1})$.

A sufficient condition for \eqref{app:Q-decay-cond} can be derived as follows.
For notational convenience, let $c_i = u_i^Tv^*$ for $i=1,2,3,4,5$.
Then, since $\sum_{i=1}^5 c_i^2 = 1$, we have
\begin{align*}
|\left\langle \text{Loss}(t), \Pi_{V_{t_s}^{\text{var}}}[v^*]\right\rangle|
\le \sum_{i=1}^4 |c_i||\langle \text{Loss}(t), u_i \rangle|
\le \|\Pi_{V_{t_s}^\text{var}} \text{Loss}(t) \|,
\end{align*}
where the Cauchy-Schwartz inequality is applied on the second inequality.
It then follows from Theorem~\ref{thm:conti-loss-n} that 
$$
\|\Pi_{V_{t_s}^\text{var}} \text{Loss}(t) \|^2
\le \|\Pi_{V_{t_s}} \text{Loss}(t) \|^2 \le \|\Pi_{V_{t_s}} \text{Loss}(t_s) \|^2 e^{-\int_{t_s}^t \bm{r}_{\min}(u)du}.
$$
Therefore, a sufficient condition for \eqref{app:Q-decay-cond}  is
\begin{align*}
e^{-\frac{1}{2}\int_{t_s}^t \bm{r}_{\min}(u)du}
< \frac{(1-|c_5|)|\langle \text{Loss}(t_s), u_5 \rangle|}{\|\Pi_{V_{t_s}} \text{Loss}(t_s) \|}.
\end{align*}

\section{Explicit Construction of Fully Trained Networks} \label{app:explicit-FTN}

\begin{proposition} \label{thm:fullyNN}
	For a partition of an interval of size $N$, which contains all the training data points $\{x_k\}_{k=1}^m$, 
	there exists a shallow ReLU network of width at most $2N$
	whose parameter is a stationary point of the gradient flow dynamics \eqref{def:bias-gradflow}.
\end{proposition}
\begin{proof}
	Let $\{I_j \}_{j=1}^N$ be a set of subintervals of the partition.
	Without loss of generality, let us assume that each $j$, $I_j \cap \{x_k\}_{k=1}^m \ne \emptyset$.
	Let
	$\mathcal{T}_m = \{(x_k,y_k)\}_{k=1}^m$ be the training data set.
	For $j \ge 1$, let 
	$$
	T_j = \left\{(x,\hat{y}) | x \in I_j, \hat{y} = y - \sum_{l=1}^{j-1} f_{l}(x), (x,y) \in \mathcal{T}_m\right\},
	$$
	with the convention of $\sum_{l=1}^0 f_l(x) = 0$.
	Let $f_j(x):=a_j^*x + d_j^*$ be the least squares (or least norm) regression line 
	obtained from $T_j$.
	
	At each $j$, by letting $z_j$ be the smallest data point in $T_j$
	and $\epsilon_j$ be the half of the distance between $z_j$ 
	and the largest data point in $T_{j-1}$ (when $j=1$, $\epsilon_1$ can be chosen any positive number),
	it follows from Lemma~\ref{lem:line-twoNeurons}
	that $f_j(x)\cdot \mathbb{I}_{z_j \ge 0}(x)$ can be exactly represented by using at most two neurons
	in $R_j = (-\infty,z_j-\epsilon_j]\cup [z_j,\infty)$.
	Since there is no data point in $(z_j - \epsilon_j, z_j)$, 
	$f^*(x) = \sum_{j=1}^N f_j(x)\cdot \mathbb{I}_{z_j \ge 0}(x)$
	can be represented by a neural network $f^{NN}(x)$ of width at most $2N$
	in $R = \cap_{j=1}^N R_j$.
	Since $R^c \cap \{x_k\}_{k=1}^m = \emptyset$, 
	$f^{NN}(x)$ satisfies the stationary conditions in Lemma~\ref{lemma:critical}.
	\begin{lemma} \label{lem:line-twoNeurons}
		For $z, a, d \in \mathbb{R}$ and $a \ne 0$, let
		\begin{equation*}
			f_z(x) = (ax + d)\cdot \mathbb{I}_{x \ge z}(x).
		\end{equation*}		
		Then, for any $\epsilon > 0$ and $\forall x \in (-\infty,z-\epsilon]\cup [z,\infty)$,
		\begin{equation}
		f(x) = \begin{cases}
		a\phi(x + \frac{d}{a}) & \text{if } 0 \le z + \frac{d}{a} < \epsilon, \\
		\left(\frac{az + b}{\epsilon}\right)\phi(x - z + \epsilon) - \left(\frac{a(z-\epsilon) + b}{\epsilon}\right)\phi(x - z) & \text{otherwise}. 
		\end{cases}
		\end{equation}
	\end{lemma}
	The proof of Lemma is straightforward. 
\end{proof}

\section{Proof of Theorem~\ref{thm:LSQ}} \label{app:thm:LSQ}
\begin{proof}
	From the COD of $A$, we observe that 
	\begin{align*}
	Q^T(A + V_l)Z 
	&= Q^TAZ + Q^TV_lZ = \hat{T}
	+
	(Q^Tv_l)\otimes Z_{l,:},
	\end{align*}
	where $\otimes$ is the outer product of vectors.
	Let 
	\begin{equation*}
	b := Q^Ty = \begin{bmatrix}
	b^{(1)} \\ b^{(2)}
	\end{bmatrix},
	\qquad
	q := Q^Tv_l = \begin{bmatrix}
	q^{(1)} \\ q^{(2)}
	\end{bmatrix}
	\qquad
	x := Z^Tc = \begin{bmatrix}
	x^{(1)} \\ x^{(2)}
	\end{bmatrix},
	\end{equation*}
	where $b^{(1)}, q^{(1)}, x^{(1)} \in \mathbb{R}^r$.
	Also, let 
	$
	Z = \begin{bmatrix}
	Z_r & \tilde{Z}
	\end{bmatrix},
	$
	where 
	$Z_r \in \mathbb{R}^{n\times r}$.
	Then, we have $c = Z_rx^{(1)} + \tilde{Z}x^{(2)}$.
	
	With the above notation, it can be checked that 
	\begin{align*}
	\|\tilde{A}c - y\|^2 
	&= \|(\hat{T} + (Q^Tv_l)\otimes Z_{l,:})x - b\|^2
	= \|\hat{T}x - b + c_l(Q^Tv_l)\|^2,
	\end{align*}
	where the last equality uses the fact that 
	$
	Z_{l,:}x = Z_{l,:}Z^Tc = e_l^Tc = c_l
	$.
	Here $e_l$ is the vector with a 1 in the $l$-th entry and 0's elsewhere,
	and the third equality holds from the mutual orthogonality of columns of $Z$.
	Thus, we have 
	\begin{equation} \label{pf:thm:lsq-eqn1}
	\begin{split}
	\|\tilde{A}c - y\|^2 
	&= \|\hat{T}w - b + c_l(Q^Tv_l)\|^2 \\
	&= \|Tx^{(1)} - b^{(1)} + c_lq^{(1)}\|^2 + \|c_lq^{(2)} - b^{(2)}\|^2.
	\end{split}
	\end{equation}
	Note that assuming $\|q^{(2)}\| > 0$, the second term in the right hand side is minimized at 
	$c_l = d_l^*=\frac{\langle b^{(2)}, q^{(2)} \rangle}{\|q^{(2)}\|^2}$,
	which can be checked as follows:
	\begin{align*}
	\|c_lq^{(2)} - b^{(2)}\|^2
	&= \|q^{(2)}\|^2c_l^2 - 2\langle b^{(2)}, q^{(2)} \rangle c_l + \|b^{(2)}\|^2
	\\
	&= \|q^{(2)}\|^2\left(c_l - d_l^* \right)^2 + \|b^{(2)}\|^2 - \frac{|\langle b^{(2)}, q^{(2)} \rangle|^2}{\|q^{(2)}\|^2}.
	\end{align*}
	If $\|q^{(2)}\| = 0$, the second term is $\|b^{(2)}\|^2$, which is independent of $c$.
	Therefore, if one finds a vector $c^*$ whose $l$-th entry is $d_l^*$
	and that makes the first term in the right hand side of \eqref{pf:thm:lsq-eqn1},
	it should be the desired solution.

	Assuming $\|\tilde{Z}_{l,:}\| > 0$, let $c^* = Zx^*$ where
	\begin{equation*}
	x^* = \begin{bmatrix}
	x_*^{(1)} \\ x_*^{(2)}
	\end{bmatrix},
	\quad 
	x_*^{(1)} = T^{-1}(b^{(1)} - d_l^*q^{(1)}), \quad
	x_*^{(2)} = \frac{d_l^* - (Z_r)_{l,:}x_*^{(1)}}{\|\tilde{Z}_{l,:}\|^2}(\tilde{Z})_{l,:}^T.
	\end{equation*}
	Here we set $d_l^* = (Z_r)_{l,:}x_*^{(1)}$ if $\|q^{(2)}\| = 0$.
	It then can be checked that 
	\begin{align*}
	[c^*]_l &= (Z_r)_{l,:}x_*^{(1)} + (\tilde{Z})_{l,:}x_*^{(2)} \\
	&=(Z_r)_{l,:}x_*^{(1)} + \frac{d_l^* - (Z_r)_{l,:}x_*^{(1)}}{\|\tilde{Z}_{l,:}\|^2}(\tilde{Z})_{l,:}(\tilde{Z})_{l,:}^T = d_l^*,
	\end{align*}
	and
	\begin{align*}
	\|Tx^{(1)}_* - b^{(1)} + d_l^*q^{(1)}\|^2
	&= \|(b^{(1)}- d_l^*q^{(1)}) - b^{(1)} + d_l^*q^{(1)}\|^2 
	= \|(d_l^* -  d_l^*)q^{(1)}\|^2 = 0.
	\end{align*}
	Since $x^{(1)}_*$ must satisfy $Tx_*^{(1)} = b^{(1)} - d_l^*q^{(1)}$,
	the degree of freedom is on the choice of $x^{(2)}$.
	Thus, we consider
	\begin{align*}
	\min \|x^{(2)}\| \quad \text{subject to } \tilde{Z}_{l,:}x^{(2)} = d_l^* - (Z_r)_{l,:}x_*^{(1)},
	\end{align*}
	and it can be checked that $x_*^{(2)}$ is the solution.
	Therefore, $c^*$ is the desired solution.
	
	Suppose $\|\tilde{Z}_{l,:}\| = 0$.
	Then, $[c]_l = (Z_r)_{l,:}x^{(1)}$. 
	Thus, \eqref{pf:thm:lsq-eqn1} becomes
	\begin{align*}
	\|\tilde{A}c - y\|^2 
	&= \|Tx^{(1)} - b^{(1)} + (Z_r)_{l,:}x^{(1)}q^{(1)}\|^2 + \|(Z_r)_{l,:}x^{(1)}q^{(2)} - b^{(2)}\|^2.
	\end{align*}
	By solving for $x^{(1)}$, the desired solution is given by $c^* = Z_rx_*^{(1)}$ where
	$x^{(1)}_* = M_l^{\dagger}s_l$,
	and $\tilde{T}_l = T + q_l^{(1)}(Z_r)_{l,:}$,
	$M_l = \tilde{T}_l^T\tilde{T}_l + \|q_l^{(2)}\|^2(Z_r)_{l,:}^T(Z_r)_{l,:}$,
	$s_l = T^T b^{(1)} + \langle y, v_l\rangle (Z_r)_{l,:}^T$.
	Its corresponding loss is $
	\|\tilde{A}c^* - y\|^2 = \|b^{(2)}\|^2 + \|b^{(1)}\|^2 - s_l^Tx^{(1)}_*$.
\end{proof}


\bibliographystyle{siamplain}
\bibliography{references}
\end{document}